\documentclass[preprint]{elsarticle}
\pdfoutput=1
\usepackage{lineno,hyperref}
\usepackage{amssymb}
\usepackage{amsmath}
\usepackage{multirow}
\usepackage{subcaption}
\modulolinenumbers[5]

\journal{Neurocomputing}

\newdefinition{definition}{Definition}

\def\equationautorefname~#1\null{%
  Equation~(#1)\null
}

\begin{document}

\begin{frontmatter}

\title{3D Human Pose Estimation with Siamese Equivariant Embedding}

\author[elte]{M\'arton V\'eges\corref{mycorrespondingauthor}}
\cortext[mycorrespondingauthor]{Corresponding author}
\ead{vegesm@caesar.elte.hu}

\author[elte]{Viktor Varga}
\author[elte]{Andr\'as L\H{o}rincz}

\address[elte]{E\"otv\"os Lor\'and University, Budapest, Hungary}

\begin{abstract}
In monocular 3D human pose estimation a common setup is to first detect 2D positions and then lift the detection into 3D coordinates. Many algorithms suffer from overfitting to camera positions in the training set. We propose a siamese architecture that learns a rotation equivariant hidden representation to reduce the need for data augmentation. Our method is evaluated on multiple databases with different base networks and shows a consistent improvement of error metrics. It achieves state-of-the-art cross-camera error rate among algorithms that use estimated 2D joint coordinates only. 
\end{abstract}


\end{frontmatter}


\section{Introduction}
Estimating human 3D poses from still images has received an increase of interest lately. The problem has many important potential applications, such as activity recognition, interaction analysis between people (e.g. object passing) and surveillance. Having the 3D coordinates of the human skeleton also helps in augmented reality applications or remote sensing.

However, the task is harder than traditional 2D pose estimation due to some fundamental differences. First, the problem formulation is inherently ambiguous: during the perspective projection information is lost and can not be retrieved. It is impossible to tell the difference between a close, short person and a tall, far away one. Second, it is difficult to create 3D pose datasets, especially in the wild. While special equipment exists to capture the position of markers attached to the body, it restricts the recordings to lab environments. The problem is aggravated by the fact that deep learning networks are data hungry and need large amounts of training examples to be robust against variations in lighting, actor appearance and background change.

One approach to solve the latter issue is to take advantage of the abundance of 2D pose annotated data by using an off-the-shelf 2D pose estimator. State-of-the-art 2D pose estimators \cite{openpose,stacked_hourglass,alphapose} have reached superior results that enable us to employ them as standalone components. Martinez et al. \cite{3dbaseline} used a pretrained Stacked Hourglass network \cite{stacked_hourglass} to generate 2D positions and then a simple fully connected network with residual blocks to achieve state-of-the-art results. Since the network does not receive the image at all, only the 2D keypoints, this approach is robust against changes in illumination and background.

However, as identified by Fang et al. \cite{fang2018posegrammar}, the above algorithm overfits to existing camera angles and does not generalize well to unseen positions. In the standard evaluation protocol of the popular Human3.6M dataset \cite{h36m}, all cameras are included both in the training and test set. When excluding one of the four cameras from the training set and restricting the test set to that camera only, the error increases significantly. Augmenting the dataset by rotating existing poses helps but only to an extent. The error is still higher compared to the original protocol even after augmentation.

To alleviate this problem, we propose a siamese network \cite{siameseSignature} based architecture that learns an equivariant embedding stable to rotations. The equivariant hidden representation has the property that applying a rotation on the input rotates the embedding the same way. This reduces the need for artificial data augmentation as some of it is already baked into the network. The siamese architecture makes it easy to teach the equivariance to the network and circumvents the need for an autoencoder. Using an autoencoder for this task has the downside that it has to learn to recreate a random noise to generate a rotated output (see Section 3.2 for detailed explanation).

Our contribution can be summarized as follows: We introduce a siamese architecture that learns a geometrically interpretable embedding. The embedding is rotationally equivariant that makes the network robust to new camera views. The architecture is tested on multiple datasets and with different base networks. We achieve state-of-the-art results on unseen camera poses on the Human3.6m dataset \cite{h36m} among methods that do not use image input directly. We also make our code publicly available\footnote{\url{https://github.com/vegesm/siamese-pose-estimation}}.

The structure of the paper is the following: in Section 2 we review the literature, in Section 3 we introduce equivariance and in Section 4 the network architecture is detailed. The performed experiments  and their results can be found in Section 5. Finally, we summarize our findings in Section 6.

\section{Related Work}
\subsection{3D Pose Estimation} 

Previous approaches focused on predicting the 3D pose directly from an image, in an end-to-end fashion. For example, in \cite{gorog}, the authors predict a 3D heatmap, gradually refining it along the depth dimension, increasing the resolution step-by-step. Zhou et al \cite{zhou2017} places a 3D regression network on top of a 2D pose estimator and extends the network with a semi-supervised loss allowing the usage of images with only 2D annotations for training. Another approach uses bone representation instead of joint coordinates \cite{Sun2017compositional}.

Compared to the above methods, Martinez et al. \cite{3dbaseline} use a 2D pose estimator and 2D pose to 3D pose regressor as separate components. The 3D regressor is a 6-layer fully connected neural network using standard techniques only, such as batch normalization or residual connections. With this simple architecture, they achieved state-of-the-art results at the time. This simplicity inspired new research expanding on the 2D to 3D pose estimator capabilities. Hossain et al. \cite{Hossain2017temporal} used temporal information by adding recurrence to the network. Fang et al. \cite{fang2018posegrammar} added bi-directional RNNs to learn additional constraints, such as symmetry or bone structure. 

Another direction of research aims to combine heatmap based approaches used extensively in 2D pose estimation \cite{openpose,stacked_hourglass} with regression based approaches used in 3D pose estimation. In \cite{integralPose}, the authors connect a 2D pose estimator generating joint location heatmaps and the 3D regression network with the soft-argmax function. The soft-argmax is a differentiable approximation of argmax whose derivative is not everywhere zero, thus the network becomes end-to-end trainable. Luvizon et. al. \cite{Luvizon2018softargmax} similarly use the soft-argmax function in a multitask estimation network.

Recently, many works included the estimation of pairwise depth rankings of joints, where the relative distance of two joints from the camera is predicted. The motivation behind the method is that it is easy for humans to annotate 2D images with depth rankings thus existing 2D pose datasets \cite{mpii-hp,lsp} can be extended and used as auxiliary training data. In \cite{pavlakos2018ordinal}, depth ranking was added to the MPII-HP and LSP datasets. The method uses these two datasets for additional weak supervision. Shi et al. \cite{fbipose} do not require the full ranking of all joints, only the bones. Wang et al. \cite{drpose}  predicts a pairwise depth ranking matrix from the image and then fuses the predicted matrix with the 2D joint location heatmaps. Their method does not use any of the extended 2D datasets. Finally, in \cite{ronchi2018allrelative} the authors analyze the performance of the human annotators on this task.

\subsection{Siamese networks}
Unlike traditional deep networks, siamese networks have two identical bran-ches sharing the same weights. Instead of a single input image, pairs of images are fed to the network and the loss is computed on the difference of the output of the two branches. Since the branches share the same weights, they are updated the same way during backpropagation and remain identical through training. Thus, in inference time, it is enough to use only one of the branches.

Siamese networks were originally proposed to solve handwriting verification \cite{siameseSignature}. Since then, it was widely used in face verification \cite{facenet,deepid2}. Siamese regression methods were also used in 3D object pose estimation. Doumanoglou et al. \cite{doumanoglouSiamesePose} used a loss that ensures that the distribution of hidden representations in the feature space is similar to that of the target datapoints in the pose space. Unlike us, they do not use an equivariant embedding on the hidden representations. In \cite{siamese_headpose}, the authors predict head poses using a siamese architecture. Compared to our work, they only have a siamese loss on the last output layer and not on an intermediate layer.

\subsection{Equivariant networks}
Equivariant networks have the promise to achieve similar performance to standard deep networks with smaller capacity and less data augmentation. In \cite{worrall2016harmonic} a new state-of-the-art is achieved on rotated-MNIST \cite{rotatedMNIST} while reaching its maximum performance using less data than a standard CNN. In \cite{cohen2018spherical}, the authors extend the standard CNNs to spheres, providing equivariance over three dimensional rotations. That formulation produces an output on the space of transformations, while the method of Esteves et al. \cite{esteves2018so3equivariance} has the sphere as an output. The latter also achieved results comparable to or better than the state-of-the-art while using a much smaller network on the ModelNet40 \cite{modelnet40} and SHREC'17 \cite{shrec17} datasets. In pose estimation, Rhodin et al. \cite{helge_geometry-aware} used an equivariant network to create an autoencoder to generate images of human poses.

\section{Background}
For completeness, we introduce equivariance \cite{worall2017interpretable}, and its weaker version, rotational equivariance.
\begin{definition}[Equivariant function] Let $f:\mathcal{X}\rightarrow\mathcal{Y}$ be a function and $T_{\theta}:\mathcal{X}\rightarrow\mathcal{X}$ and $U_{\theta}:\mathcal{Y}\rightarrow\mathcal{Y}$ two sets of transformations parametrized by $\theta$. We say $f$ is equivariant to $T$ and $U$ if
$$f(T_{\theta}(x))=U_{\theta}(f(x))$$
for all $x\in\mathcal{X}$ and $\theta$.
\end{definition}
What this means is that $T_{\theta}$ and $U_{\theta}$ are a pair of transformations whose order with $f$  can be exchanged upon replacing one with the other.
That is, transforming the input with $T$ and then applying $f$ is the same as first applying $f$ and then transforming the output with $U$. If a neural network is equivariant, the network will automatically learn to be robust against transformations in $T$. This way less augmentation is needed as the augmentation transformations are already handled by the network. Typical examples are fully convolutional networks. They are translation-equivariant and during training usually no translation augmentations are applied, just rotations and reflections.

Now, we move on to rotational equivariance, defined in \cite{helge_geometry-aware}. The definition  below is specific to how equivariance is used in our algorithm. First note that following \cite{3dbaseline} we split the task into two steps: first, predicting the 2D pose $P_{2D}\in\mathbb{R}^{2\times n}$ from the input image, then predicting the 3D pose $P_{3D}\in\mathbb{R}^{3\times n}$ from $P_{2D}$ only where $n$ is the number of joints. In the second step no image information was used, just the coordinates of the 2D skeleton.

\begin{definition}[Rotational equivariance] 
Let the hidden representation $h$ be a set of $M$ 3-dimensional vectors (i.e. $h\in\mathbb{R}^{3\times M}$) and $f:\mathbb{R}^{2\times n}\rightarrow\mathbb{R}^{3\times M}$ be an encoder that takes the input 2D position into the hidden representation $h$, that is $f(P_{2D})=h$. $f$ is rotationally equivariant, if:
\begin{equation} \label{eq:equiv}
    f(\mathrm{\Pi} \left(RP_{3D}\right))=Rf(\mathrm{\Pi} P_{3D}),
\end{equation}
where $\mathrm{\Pi}$ is the 3D to 2D projection and $R\in\mathbb{R}^{3\times3}$ is a rotation matrix.
\end{definition}

In other words, rotating the input pose and applying the encoder $f$ has the same effect as encoding the pose and then rotating the hidden representation. Equivalently, the order of the rotation and the encoder can be swapped. 

\section{Method}
As mentioned in the previous section, the 3D pose estimation is performed in two steps: first the 2D pose is determined with an off-the-shelf component and then the 3D position is predicted from the 2D skeleton. We focus on the second step here.

The goal is to create a network that is robust against unseen camera angles without excessive augmentation. Note that seeing a pose P from a new (rotated) camera angle is equivalent to seeing that same pose from a fixed angle but the pose itself rotated the other direction. So we can rephrase our goal as being robust against unseen rotations of a pose. To achieve this, we would like our network to learn a hidden representation $h$ that is rotationally equivariant to the input.

To learn equivariance, it is possible to use an autoencoder with dynamically rotating the hidden representation during training \cite{helge_geometry-aware}. However, our inputs are noisy 2D pose estimations from another detector, thus an autoencoder would have to learn to simulate the prediction error of the 2D pose estimator. Instead we are opting to use a siamese architecture, which has the advantage that it does not have to learn a complete encoding of the input, contrary to an autoencoder. This makes further extension of the model to image inputs much easier as only information needed for the pose estimation must be encoded in the hidden representation. 

\begin{figure}[ht]
    \centering
    \includegraphics[width=0.8\textwidth]{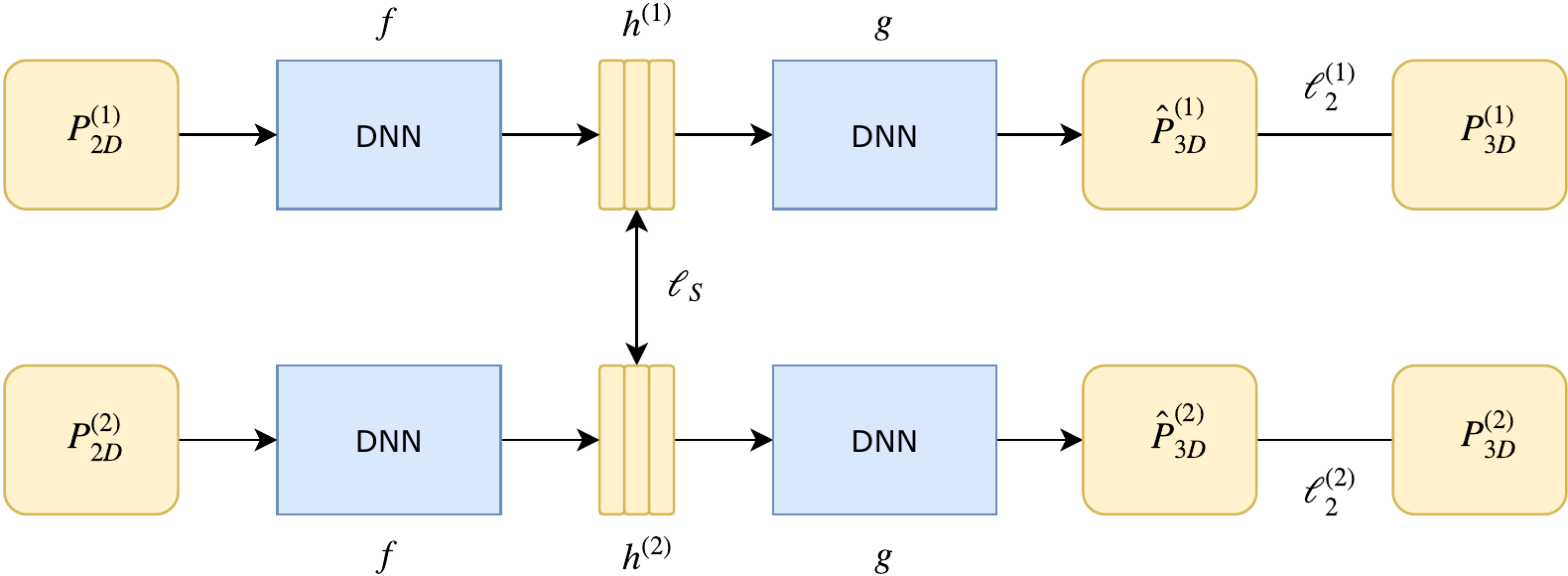}
    \caption{\textbf{Our siamese architecture.} We feed the input 2D detections $P_{2D}^{(1)}$ and $P_{2D}^{(2)}$ to the two branches of the network. The encoder network $f$ converts them into hidden representations $h^{(1)},h^{(2)}\in\mathbb{R}^{3\times M}$. Afterwards, the decoder network $g$ converts $h^{(i)}$ into the final 3D predictions $\hat{P}_{3D}^{(i)}$. Both outputs have an $L_2$ loss applied on them. We also apply an additional siamese loss function $\ell_S$ based on the hidden representations $h^{(i)}$.}
    \label{fig:architecture}
\end{figure}

A high level overview of our network is presented on \autoref{fig:architecture}. It has two identical branches split into an encoder $f$ and decoder $g$. \autoref{eq:equiv} is enforced by a siamese loss described in the next section. 

\subsection{Equivariant siamese loss}
Assume we have calibrated cameras and know their rotation matrices relative to a suitable 3D coordinate system. Let $C_1$ and $C_2$ be two cameras, and the rotation matrix taking the view of $C_1$ into $C_2$ to be $R$. Let $P_{3D}^{(1)}$ and $P_{3D}^{(2)}$ the same pose in the first and second camera coordinate system respectively and denote its hidden representations under the two cameras with $h_1$ and $h_2$. Using~(\ref{eq:equiv}) and the fact that $h_i=f\left(\mathrm{\Pi} P_{3D}^{(i)}\right)$:
\begin{equation}
\begin{split}
\label{eq:hidden}
    Rh_1&=Rf\left(\mathrm{\Pi} \left(P_{3D}^{(1)}\right)\right)= \\
   & =f\left(\mathrm{\Pi} \left(RP_{3D}^{(1)}\right)\right)=f\left(\mathrm{\Pi} \left(P_{3D}^{(2)}\right)\right)=h_2.
\end{split}
\end{equation}

The equation above can be enforced by a siamese network naturally. If the input poses are $P_{2D}^{(1)}=\mathrm{\Pi} \left(P_{3D}^{(1)}\right)$ and $P_{2D}^{(2)}=\mathrm{\Pi} \left(P_{3D}^{(2)}\right)$ then adding a loss on $\left\Vert Rh_1-h_2\right\Vert$ forces the network to optimize for \autoref{eq:hidden}. 

However, this would only work for input pairs where the two inputs represent the same pose from different angles. To allow inputs representing different poses, first assume that there is some canonical coordinate system and $R_1$ and $R_2$ are the rotation matrices going from this absolute system to one relative to $C_1$ and $C_2$, respectively. Let $P_{3DA}^{(1)}$ be the pose in this absolute coordinate system supplied to the first camera and $P_{3DA}^{(2)}$ supplied to the second camera. Then the 2D inputs for the network are denoted by $P_{2D}^{(1)}=\mathrm{\Pi} \left(R_1P_{3DA}^{(1)}\right)$ and $P_{2D}^{(2)}=\mathrm{\Pi} \left(R_2P_{3DA}^{(2)}\right)$. Thus:
\begin{equation*}
\begin{split}
    R_2R_1^{-1}h_1&=R_2R_1^{-1}f\left(\mathrm{\Pi} \left(R_1P_{3DA}^{(1)}\right)\right)= \\
   & =f\left(\mathrm{\Pi} \left(R_2R_1^{-1}R_1P_{3DA}^{(1)}\right)\right)=f\left(\mathrm{\Pi} \left(R_2P_{3DA}^{(1)}\right)\right).
\end{split}
\end{equation*}
Since $h_2=f\left(\mathrm{\Pi} \left(R_2P_{3DA}^{(2)}\right)\right)$ by definition,  it is reasonable to have 
$$\left\Vert R_2R_1^{-1}h_1-h_2\right\Vert\approx\lambda_1\left\Vert R_2P_{3DA}^{(1)}-R_2P_{3DA}^{(2)}\right\Vert=\lambda_1\left\Vert P_{3DA}^{(1)}-P_{3DA}^{(2)}\right\Vert,$$
where $\lambda_1$ is a scaling parameter. In the second equality we used the fact that $R_2$ is a rotation matrix thus orthonormal. Now we can formulate the loss as:
\begin{equation}
\label{eq:siam_loss}
    \ell_S=\left|\left\Vert R_2R_1^{-1}h_1-h_2\right\Vert-\lambda_1\left\Vert P_{3DA}^{(1)}-P_{3DA}^{(2)}\right\Vert\right|^2.
\end{equation}
This is similar to the loss used in \cite{doumanoglouSiamesePose,siamese_headpose}.

\subsection{Network structure}
\begin{figure}[ht]
    \centering
    \includegraphics[width=0.7\textwidth]{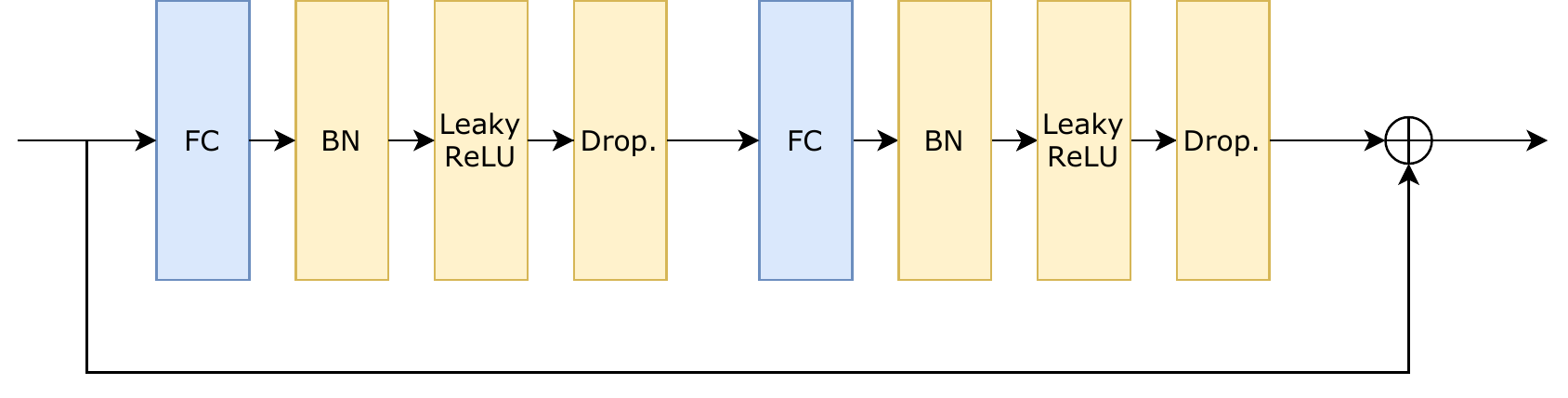}
    \caption{\textbf{Residual modules used in the network.} One residual module consists of two fully connected layers of 1024 nodes followed by batch normalization \cite{batchnorm} and dropout \cite{dropout}. The activation layer is Leaky-ReLU to solve problems with dying ReLUs.}
    \label{fig:res-block}
\end{figure}
The structure of the network is illustrated on \autoref{fig:architecture}. It has two identical branches, each branch is built up from an encoder $f$ and a decoder $g$, for which $g(f(P_{2D}))=g(h)=P_{3D}$.

The main component of both $f$ and $g$ is a single residual module, depicted on \autoref{fig:res-block}. The architecture was inspired by \cite{3dbaseline}. Each fully connected layer has 1024 nodes. The encoder $f$ has a dense layer before the residual block to scale up the input to a dimension of 1024. In the decoder, after the residual block a dense layer with 48 nodes produces the final output. We have found that dying ReLUs were a problem so used Leaky-ReLUs as activation functions instead of regular ReLUs. 

To resize the output of the encoder from 1024 to $3M$ a dense layer is used with no activation function. The resulting vector of length $3M$ is reshaped to $3\times M$ and normalized along the first axis, similarly to \cite{facenet} and \cite{amazon_siamese-sampling}. After the embedding, the output tensor is resized back to 1024 with another fully connected layer. It was found empirically that placing a batch normalization and dropout layer after this layer decreased the performance considerably so they were omitted.

Additionally to the siamese loss introduced in the previous section, we also add an $L_2$ loss on both outputs of the siamese network. Thus the total loss is the following:
$$\ell=\ell_2^{(1)}+\ell_2^{(2)}+\lambda_2\ell_S,$$
where $\ell_2^{(1)}$ and $\ell_2^{(2)}$ are the squared $L_2$ losses on the two branches and $\lambda_2$ is a hyperparameter.

\section{Experiments}
We have evaluated our method on multiple databases both qualitatively and quantitatively. Also extensive ablation studies were performed to validate each component of the network.  In this section we introduce these experiments and describe the implementation details.

\subsection{Database and Evaluation Protocols}
Currently one of the largest databases having 3D human poses is the Human3.6M dataset which contains 11 actors performing 15 different actions re\-cor\-ded from four camera angles. The standards error metric is the mean per joint position error (MPJPE) which is the average L2 error over all joints. There are three protocols, the last of them recently introduced by Fang et al. \cite{fang2018posegrammar} to measure cross-camera efficiency.

\begin{description}
\item[Protocol \#1] splits the dataset to training and test set by subjects. Subjects 1, 5, 6, 7, 8 are in the training set; subjects 9 and 11 are in the test set. The two splits share the same cameras and actions.

\item[Protocol \#2] has the same split as Protocol \#1. The difference is in the error metric. The MPJPE is calculated after an affine Procrustean alignment to the ground truth using rotations and translations. This protocol aims to evaluate the correctness of the pose relative to itself, without taking into account scaling or rotations.

\item[Protocol \#3] aims to measure how well the method generalizes to unknown camera angles. This is similar to Protocol \#1, using the same split of subjects. However, only 3 of the cameras are in the training set and the fourth one is in the test set. Like with Protocol \#1, all actions occur in both subsets. We also call Protocol \#3 cross-camera setup.
\end{description}

\autoref{tbl:h3dm-size} contains a summary of the size of the training and test set split by actions. Note that Protocol \#2 uses the same split of training set as Protocol \#1.

\begin{table}[h]
\centering
\begin{tabular}{l|rr|rr}
\multirow{2}{2.2cm}{Action} & \multicolumn{2}{c|}{Protocol \# 1} & \multicolumn{2}{c}{Protocol \# 3} \\
 & Train & Test & Train & Test \\
\hline
Directions & 100.9 & 33.7 & 75.7 & 8.4 \\
Discussion & 158.8 & 64.2 & 119.1 & 16.1 \\
Eating & 109.4 & 39.3 & 82.1 & 9.8 \\
Greeting & 72.4 & 30.6 & 54.3 & 7.7 \\
Phoning & 115.8 & 56.1 & 86.9 & 14.0 \\
Photo & 76.0 & 29.3 & 57.0 & 7.3 \\
Posing & 69.5 & 27.3 & 52.1 & 6.8 \\
Purchases & 63.1 & 19.3 & 47.3 & 4.8 \\
Sitting & 116.5 & 40.3 & 87.4 & 10.1 \\
SittingDown & 129.2 & 33.3 & 96.9 & 8.3 \\
Smoking & 133.3 & 55.6 & 99.9 & 13.9 \\
Waiting & 115.3 & 37.9 & 86.5 & 9.5 \\
WalkDog & 79.4 & 28.3 & 59.6 & 7.1 \\
WalkTogether & 87.3 & 26.2 & 65.5 & 6.5 \\
Walking & 132.7 & 29.3 & 99.6 & 7.3 \\
\hline
Total & 1559.8 & 550.6 & 1169.8 & 137.7
\end{tabular}
\caption{\textbf{Number of training examples in Human3.6M.} Numbers are in thousands.}
\label{tbl:h3dm-size}
\end{table}

\subsection{Implementation details}

\subsubsection{Preprocessing and augmentation} 
To predict the 2D pose, we use a Stacked Hourglass network \cite{stacked_hourglass} pretrained on the MPII-HP dataset \cite{mpii-hp} and fine-tuned on the Human3.6M \cite{h36m} database.

Following the standard setup, $P_{3D}$ is represented in a self-centered coordinate system. The hip is moved to the origin and the coordinate axes are parallel to the camera plane. Similarly to Martinez et al. \cite{3dbaseline}, we normalize both the 2D inputs and 3D targets by subtracting the mean and dividing with the standard deviation. 

To help training, we also generate augmented camera angles using the me\-thod described in \cite{fang2018posegrammar}. Note that we restrict ourselves to rotations around the central vertical axis only thus new cameras are generated on the circle the original cameras reside on. This is because in the Human3.6m dataset all cameras are on the same plane. Unlike \cite{fang2018posegrammar}, we synthesize a camera every 15 degrees and not 30. We have removed the two closest synthetic cameras to the test camera, as in \cite{fang2018posegrammar}. To have comparable results to previously published algorithms, we did not use the augmentation on Protocols \#1 and \#2. 

For Protocol \#3, the input data was subsampled at 10fps. This was done for two reasons: first, due to augmentation the training data is quite large and using a subset of the data speeds up training; second, it helps comparing to previous work as the same sampling was applied there.

\subsubsection{Training details} 
We used a dropout rate of 0.2. We have found empirically that it yielded better results then the standard value of 0.5. This confirms our hypothesis that the siamese loss acts as a regularizer.

For training we used the Adam optimizer with a learning rate of 0.001 and an exponential decay with a rate of 0.96. The batch size was set to 256. The training ran for 100 epochs. The siamese scaling factor was empirically set to $\lambda_1=0.01$. We have found that while changes to $\lambda_1$ larger than a magnitude affect the performance considerably, smaller changes have negligible effect.  The size of the embedding was $M=128$. Using larger $M$s did not provide better results.

The selection of the pairs fed to the network was the following: in a single batch, half of the input pairs were the same poses (the same frame of the same video sequence) from different random camera angles and half of them were randomly selected poses from random camera angles. We did not investigate the effects of other sampling techniques.

\begin{figure}[ht]
    \centering
    \includegraphics[width=0.8\textwidth]{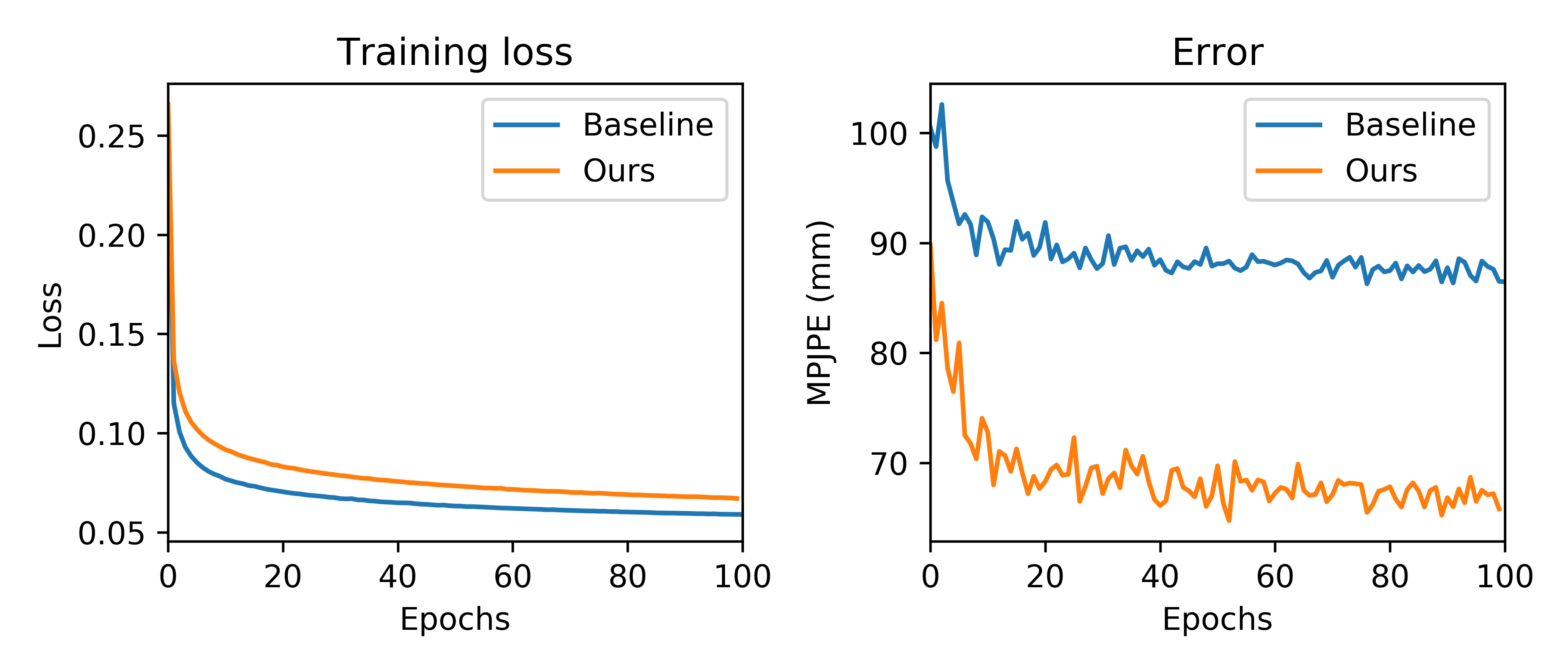}
    \caption{\textbf{Training curves.} The training loss and test error of the baseline and our model.}
    \label{fig:training-curve}
\end{figure}

To show the stability of the model we present training curves on \autoref{fig:training-curve}. Our model converges well and has a similar test error variance to the baseline algorithm. Neither the baseline nor our model overfits, the test error decreases steadily and then reaches a  plateau. Due to the exponential decay of the learning rate the test accuracy stabilizes over time.

\subsection{Quantitative results}
\begin{table}[h]
\resizebox{\textwidth}{!}{
\begin{tabular}{lcccccccccc}
\hline
Protocol \#1 & Uses Image & Direct. & Discuss & Eating & Greet & Phone & Photo & Pose & Purch. \\
\hline
LinKDE \cite{h36m} & Y & 132.7 & 183.6 & 132.3 & 164.4 & 162.1 & 205.9 & 150.6 & 171.3 \\
Zhou et al. \cite{zhou2017} & Y & 54.8 & 60.7 & 58.2 & 71.4 & \underline{62.0} & \underline{65.5} & 53.8 & 55.6 \\
DRPose3D \cite{drpose} & Y & \underline{49.2} & 55.5 & \underline{53.6} & \underline{53.4} & 63.8 & 67.7 & \underline{50.2} & \underline{51.9} \\
Martinez et al. \cite{3dbaseline} & N & 51.8 & 56.2 & 58.1 & 59.0 & 69.5 & 78.4 & 55.2 & 58.1 \\
Fang et al. \cite{fang2018posegrammar} & N & 50.1 & \underline{\textbf{54.3}} & 57.0 & 57.1 & \textbf{66.6} & \textbf{73.3} & 53.4 & 55.7 \\
\textbf{Ours} & N & \textbf{50.1} & 54.7 & \textbf{56.0} & \textbf{56.5} & 67.7 & 76.4 & \textbf{53.1} & \textbf{54.7} \\
\hline
Protocol \#1 & Uses Image & Sitting & SittingD. & Smoke & Wait & WalkD. & Walk & WalkT. & Avg. \\
\hline
LinKDE \cite{h36m} & Y & 151.6 & 243.0 & 162.1 & 170.7 & 177.1 & 96.6 & 127.9 & 162.1 \\
Zhou et al. \cite{zhou2017} & Y & 75.2 & 111.6 & 64.1 & 66.0 & \underline{51.4} & 63.2 & 55.3 & 64.9 \\
DRPose3D \cite{drpose} & Y & \underline{70.3} & \underline{81.5} & \underline{57.7} & \underline{51.5} & 58.6 & \underline{44.6} & \underline{47.2} & \underline{57.8} \\
Martinez et al. \cite{3dbaseline} & N & 74.0 & 94.6 & 62.3 & 59.1 & 65.1 & 49.5 & 52.4 & 62.9 \\
Fang et al. \cite{fang2018posegrammar} & N & \textbf{72.8} & \textbf{88.6} & \textbf{60.3} & \textbf{57.7} & \textbf{62.7} & \textbf{47.5} & 50.6 & \textbf{60.4} \\
\textbf{Ours} & N & 73.3 & 93.2 & 60.4 & 58.5 & 62.8 & 51.5 & \textbf{48.2} & 61.1 \\
\hline
\end{tabular}}
\caption{\textbf{Results on Protocol \#1.} The table shows mean joint errors in millimeters. Best results among 2D to 3D methods are selected in bold, best overall results are underlined.}
\label{tbl:results-p1}
\end{table}

\begin{table}[ht]
\resizebox{\textwidth}{!}{
\begin{tabular}{lcccccccccc}
\hline
Protocol \#2 & Uses Image & Direct. & Discuss & Eating & Greet & Phone & Photo & Pose & Purch. \\
\hline
DRPose3D \cite{drpose} & Y & \underline{36.6} & \underline{41.0} & \underline{40.8} & \underline{41.7} & \underline{45.9} & \underline{48.0} & \underline{37.0} & \underline{37.1} \\
Martinez et al. \cite{3dbaseline} & N & 39.5 & 43.2 & 46.4 & 47.0 & 51.0 & 56.0 & 41.4 & 40.6 \\
Fang et al. \cite{fang2018posegrammar} & N & \textbf{38.2} & \textbf{41.7} & \textbf{43.7} & \textbf{44.9} & \textbf{48.5} & \textbf{55.3} & \textbf{40.2} & \textbf{38.2} \\
\textbf{Ours} & N & 42.2 & 44.8 & 47.5 & 47.6 & 54.8 & 57.8 & 42.2 & 40.8 \\
\hline
Protocol \#2 & Uses Image & Sitting & SittingD. & Smoke & Wait & WalkD. & Walk & WalkT. & Avg. \\
\hline
DRPose3D \cite{drpose} & Y & \underline{51.9} & \underline{60.4} & \underline{43.9} & \underline{38.4} & \underline{42.7} & \underline{32.9} & \underline{37.2} & \underline{42.9} \\
Martinez et al. \cite{3dbaseline} & N & 56.5 & 69.4 & 49.2 & 45.0 & 49.5 & 38.0 & 43.1 & 47.7 \\
Fang et al. \cite{fang2018posegrammar} & N & \textbf{54.5} & \textbf{64.4} & \textbf{47.2} & \textbf{44.3} & \textbf{47.3} & \textbf{36.7} & 41.7 & \textbf{45.7} \\
\textbf{Ours} & N & 60.5 & 69.8 & 50.8 & 47.4 & 51.1 & 44.3 & \textbf{40.0} & 49.4 \\
\hline
\end{tabular}}
\caption{\textbf{Results on Protocol \#2.} The table shows mean joint errors in millimeters. Best results among 2D to 3D methods selected are in bold, best overall results are underlined.}
\label{tbl:results-p2}
\end{table}

\begin{table}[ht]
\resizebox{\textwidth}{!}{
\begin{tabular}{lcccccccccc}
\hline
Protocol \#3 & Uses Image & Direct. & Discuss & Eating & Greet & Phone & Photo & Pose & Purch. \\
\hline
Zhou et al. \cite{zhou2017}* & Y & 61.4 & 70.7 & 62.2 & 76.9 & 71.0 & 81.2 & 67.3 & 71.6 \\
DRPose3D \cite{drpose}$^\dagger$ & Y & 55.8 & \underline{56.1} & 59.0 & \underline{59.3} & 66.8 & \underline{70.9} & \underline{54.0} & \underline{55.0} \\
Martinez et al. \cite{3dbaseline}* & N & 65.7 & 68.8 & 92.6 & 79.9 & 84.5 & 100.4 & 72.3 & 88.2 \\
Martinez et al. \cite{3dbaseline}$^\dagger$ & N & 58.4 & 58.4 & 69.9 & 65.4 & 70.3 & 80.5 & 61.6 & 69.4 \\
Fang et al. \cite{fang2018posegrammar}$^\ddagger$ & N & 57.5 & 57.8 & 81.6 & 68.8 & 75.1 & 85.8 & 61.6 & 70.4 \\
Fang et al. \cite{fang2018posegrammar}$^\dagger$ & N & 57.8 & 57.6 & 66.3 & 65.0 & 68.4 & 79.5 & 61.8 & 67.9 \\
Ours$^\dagger$ & N & \underline{\textbf{54.5}} & \textbf{57.6} & \underline{\textbf{58.7}} & \textbf{62.3} & \underline{\textbf{66.7}} & \textbf{74.6} & \textbf{59.9} & \textbf{65.6} \\
\hline
Protocol \#3 & Uses Image & Sitting & SitingD. & Smoke & Wait & WalkD. & Walk & WalkT. & Avg. \\
\hline
Zhou et al. \cite{zhou2017}* & Y & 96.7 & 126.1 & 68.1 & 76.7 & \underline{63.3} & 72.1 & 68.9 & 75.6 \\
DRPose3D \cite{drpose}$^\dagger$ & Y & \underline{78.8} & \underline{92.4} & \underline{58.9} & \underline{56.2} & 64.6 & \underline{56.6} & \underline{55.5} & \underline{62.8} \\
Martinez et al. \cite{3dbaseline}* & N & 109.5 & 130.8 & 76.9 & 81.4 & 85.5 & 69.1 & 68.2 & 84.9 \\
Martinez et al. \cite{3dbaseline}$^\dagger$ & N & 86.8 & 99.5 & 64.5 & 69.5 & 69.6 & 60.5 & 60.2 & 69.6 \\
Fang et al. \cite{fang2018posegrammar}$^\ddagger$ & N & 95.8 & 106.9 & 68.5 & 70.4 & 73.8 & \textbf{58.5} & 59.6 & 72.8 \\
Fang et al. \cite{fang2018posegrammar}$^\dagger$ & N & 83.3 & 94.5 & 63.1 & \textbf{66.8} & \textbf{68.2} & 59.0 & \textbf{57.1} & 67.8 \\
Ours$^\dagger$ & N & \textbf{80.5} & \textbf{93.6} & \textbf{60.6} & 66.9 & 68.3 & 59.0 & 58.6 & \textbf{65.8} \\
\hline
\end{tabular}}
\caption{\textbf{Results on Protocol \#3.} The table shows mean joint errors in millimeters. Best results among 2D to 3D methods are selected in bold, best overall results are underlined. * No augmentations. $^\dagger$ Synthetic cameras every 15 degrees.  $^\ddagger$ Synthetic cameras every 30 degrees. }
\label{tbl:results-p3}
\end{table}
The results are presented in Tables \ref{tbl:results-p1}-\ref{tbl:results-p3}. In Protocol \#3, among methods using only 2D pose information and no image input, our method achieves state-of-the-art result, improving 7mm (9.6\%) over the previous best. It performs comparably to methods that use image information as well, only being 3mm (4.7\%) worse than the best method.

In Protocol \#1, our method performs better than the baseline (61.1mm vs 62.9). However, it can not beat algorithms that use image information or different network structure. This is in line with our expectations, as our extension is primarily a regularization for cross camera setup and adds little value if all the camera angles are present in both the test and training set.

\subsection{Qualitative results}
\begin{figure}
\centering
\begin{tabular}{cccccccc}
\includegraphics[width = 0.09\linewidth]{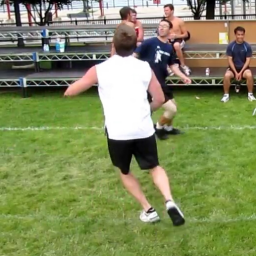} &
\includegraphics[width = 0.09\linewidth]{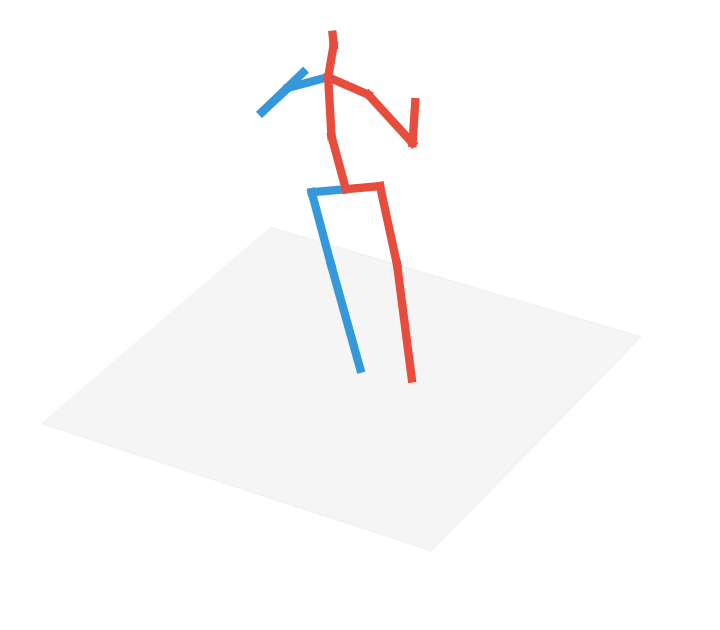} &
\includegraphics[width = 0.09\linewidth]{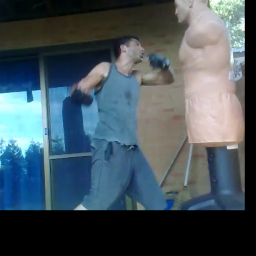} &
\includegraphics[width = 0.09\linewidth]{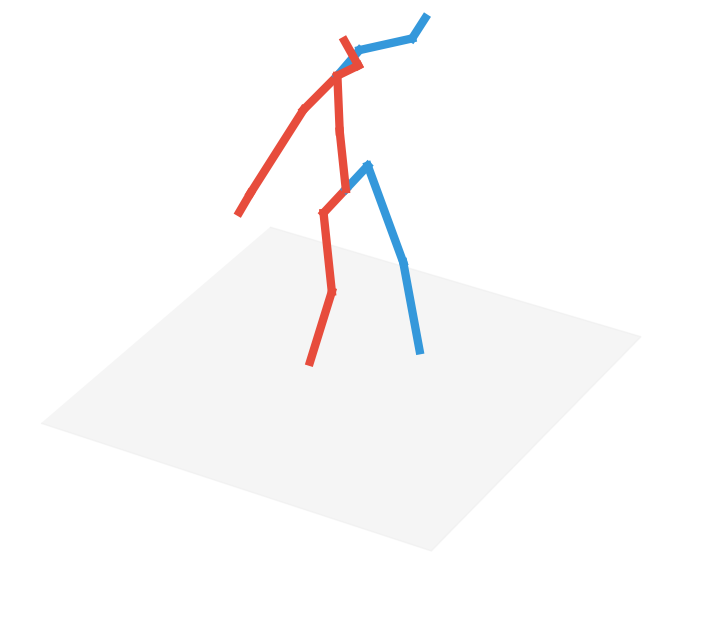} &
\includegraphics[width = 0.09\linewidth]{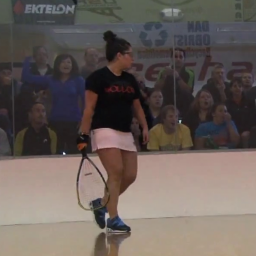} &
\includegraphics[width = 0.09\linewidth]{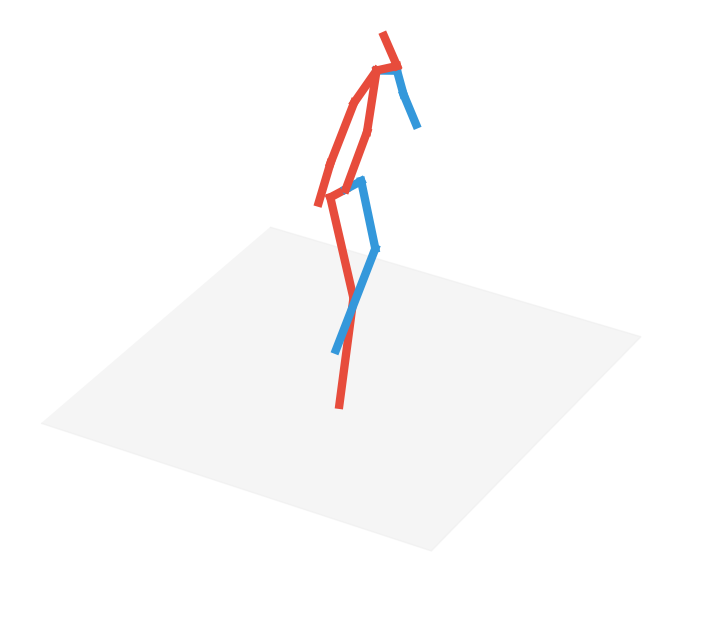} &
\includegraphics[width = 0.09\linewidth]{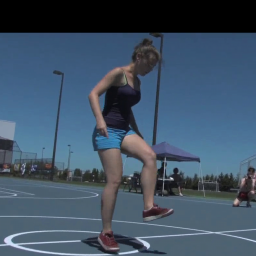} &
\includegraphics[width = 0.09\linewidth]{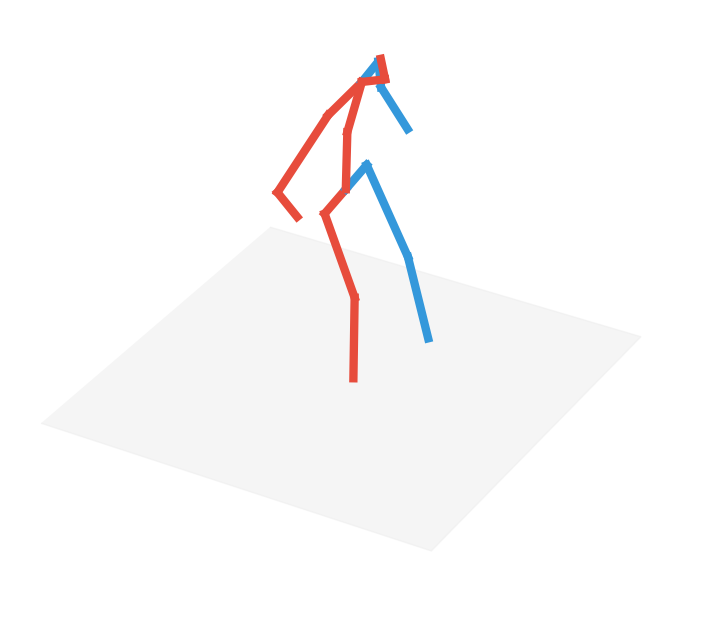} \\
\includegraphics[width = 0.09\linewidth]{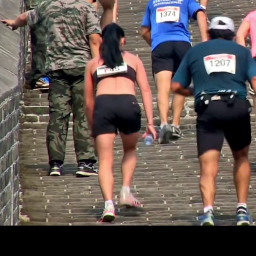} &
\includegraphics[width = 0.09\linewidth]{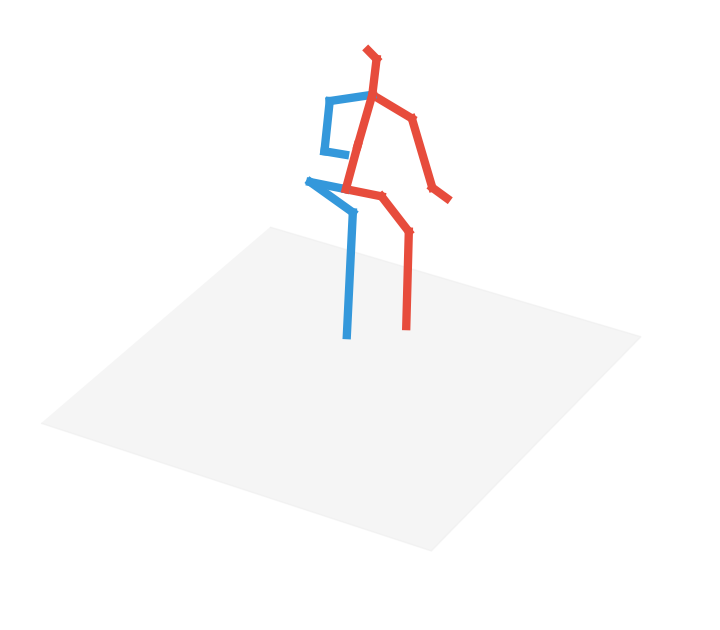} &
\includegraphics[width = 0.09\linewidth]{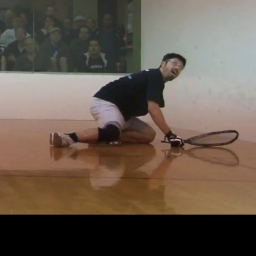} &
\includegraphics[width = 0.09\linewidth]{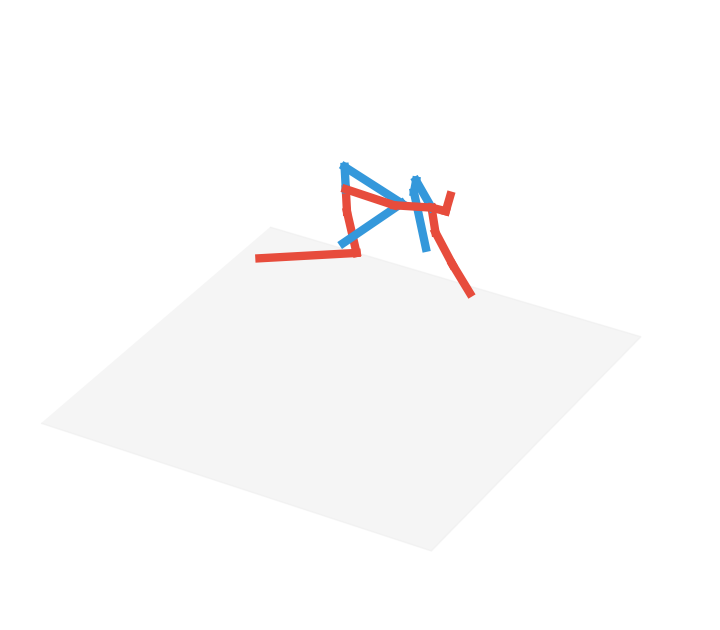} &
\includegraphics[width = 0.09\linewidth]{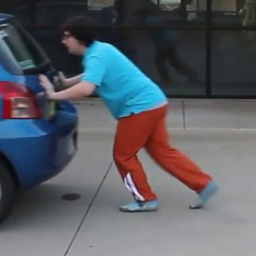} &
\includegraphics[width = 0.09\linewidth]{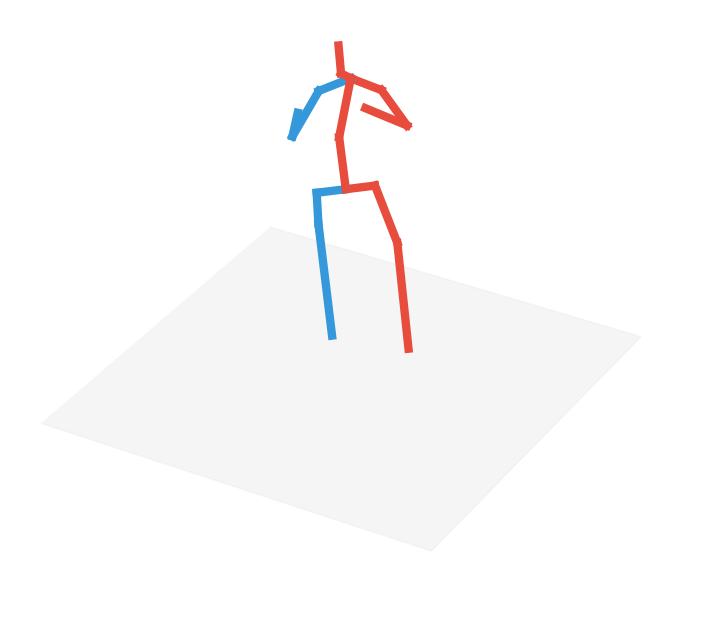} &
\includegraphics[width = 0.09\linewidth]{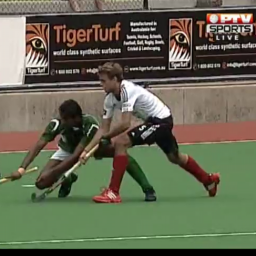} &
\includegraphics[width = 0.09\linewidth]{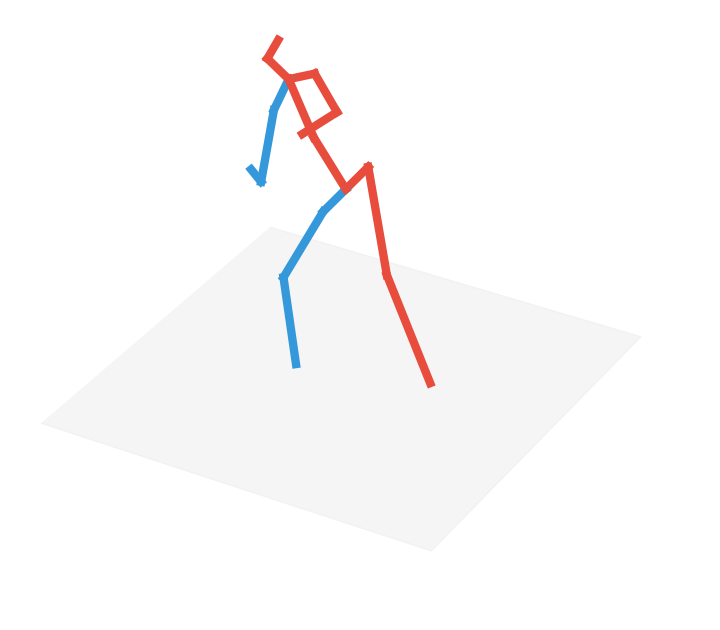} \\
\includegraphics[width = 0.09\linewidth]{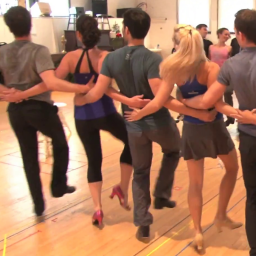} &
\includegraphics[width = 0.09\linewidth]{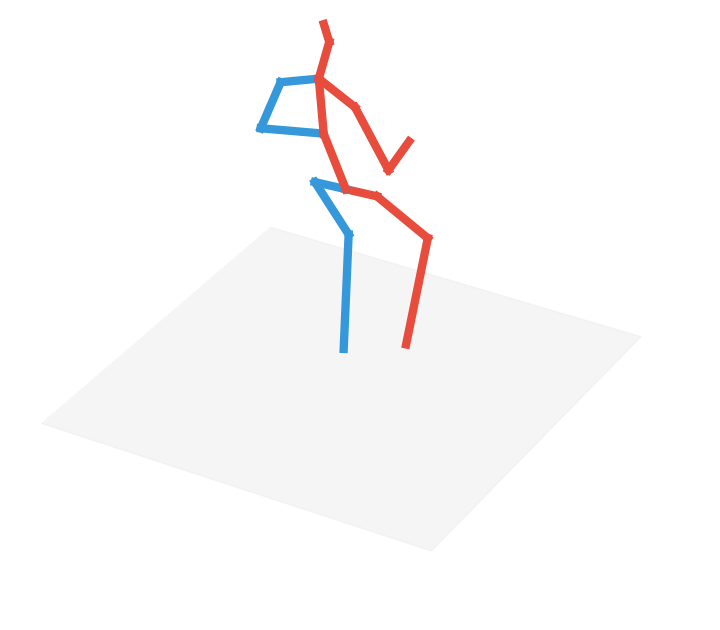} &
\includegraphics[width = 0.09\linewidth]{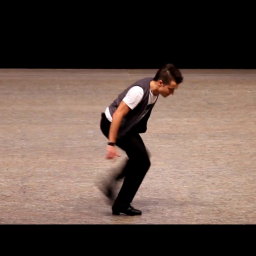} &
\includegraphics[width = 0.09\linewidth]{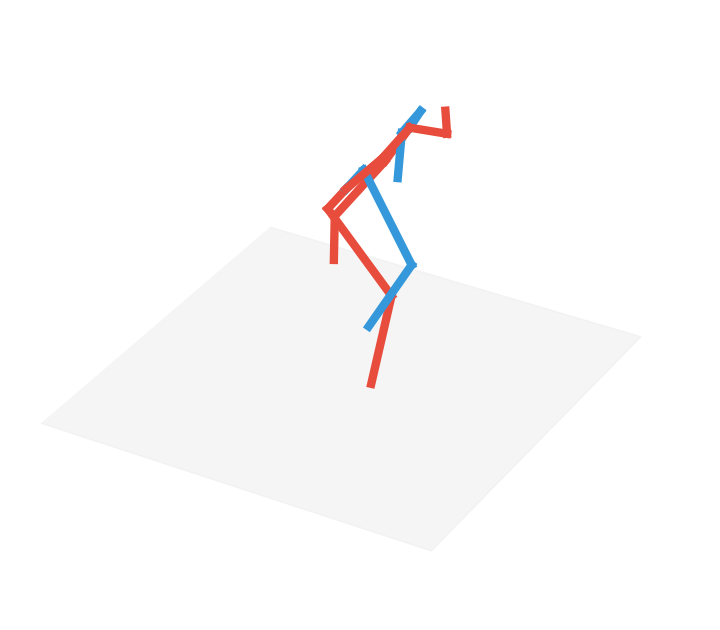} &
\includegraphics[width = 0.09\linewidth]{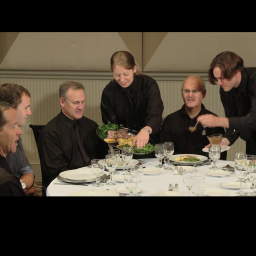} &
\includegraphics[width = 0.09\linewidth]{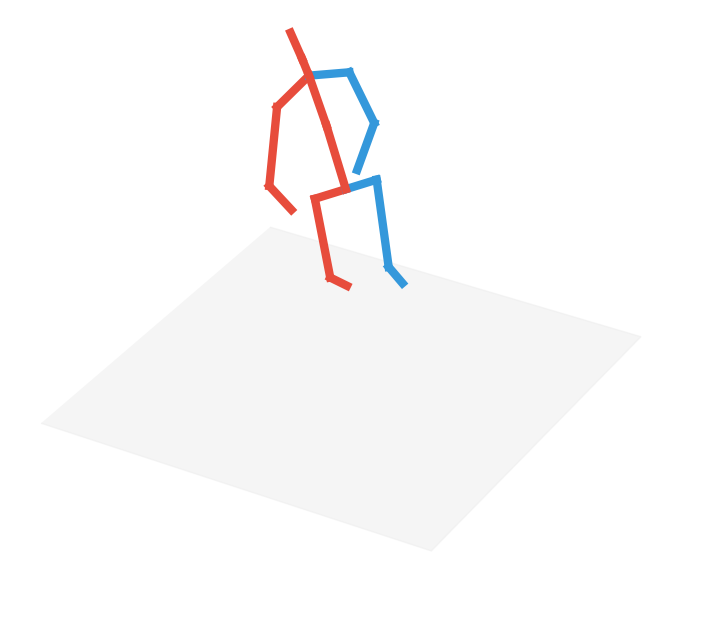} &
\includegraphics[width = 0.09\linewidth]{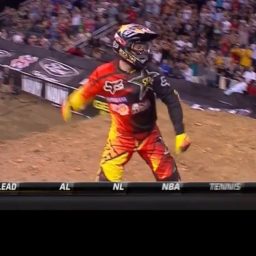} &
\includegraphics[width = 0.09\linewidth]{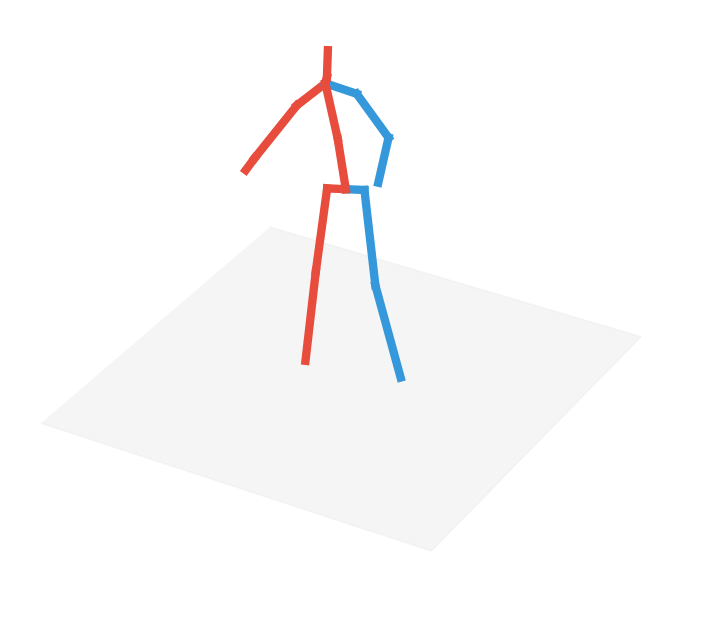} \\
\end{tabular}
\caption{\textbf{Qualitative results on MPII-2D.} The (cropped) input images are on the left and our network's prediction on the right.}
\label{fig:mpii-2d}
\end{figure}

We also show qualitative results on the MPII-HP in-the-wild dataset (\autoref{fig:mpii-2d}). For this evaluation, we trained the model on Human3.6m using protocol \#3. The MPII-HP database does not have 3D annotations so quantitative results are not available, however the presented images show that our model generalizes to new environments well. One limitation of our method is that it does not handle joints not present in the image (e.g. \autoref{fig:mpii-2d}, bottom row third image) since it was trained on images with full body poses.

\subsection{Visualizing the hidden representation}
\begin{figure}
\centering
\begin{tabular}{cccccc}
\includegraphics[width = 0.12\linewidth]{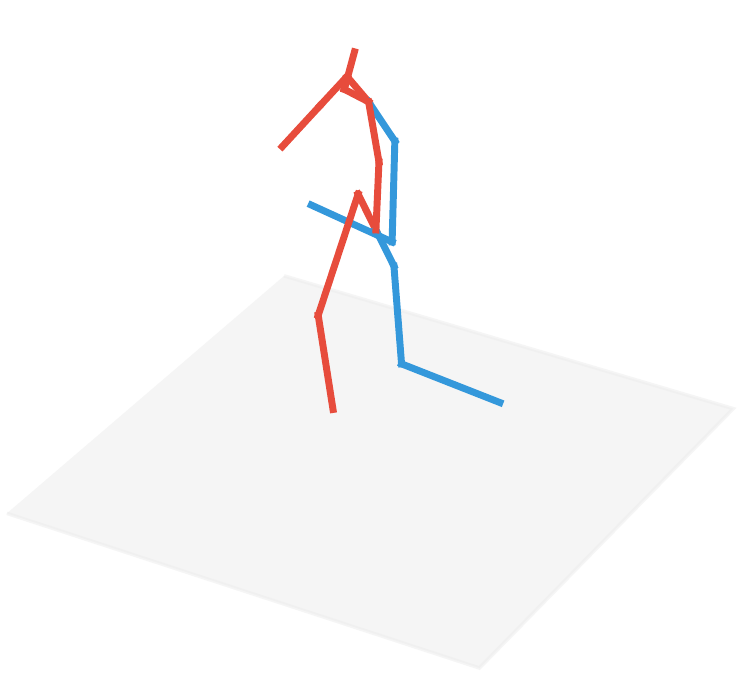} &
\includegraphics[width = 0.12\linewidth]{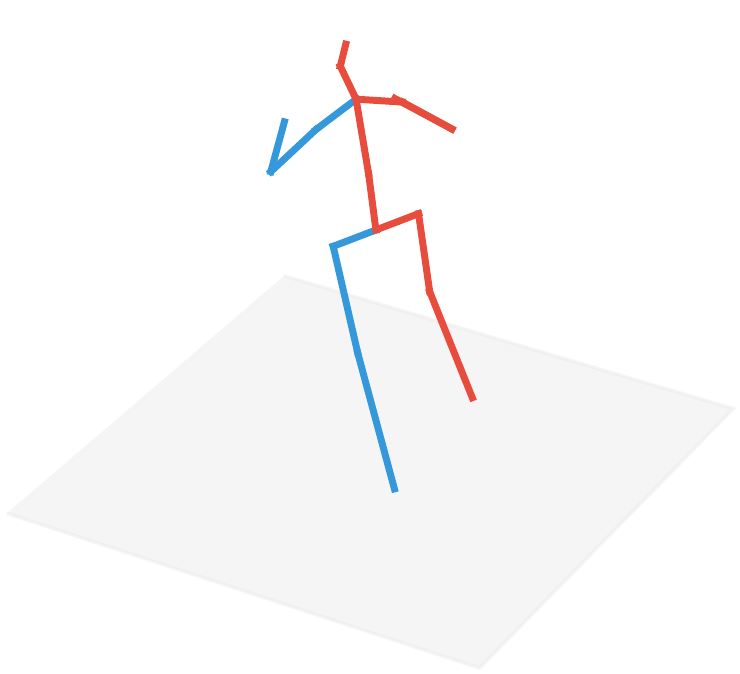} &
\includegraphics[width = 0.12\linewidth]{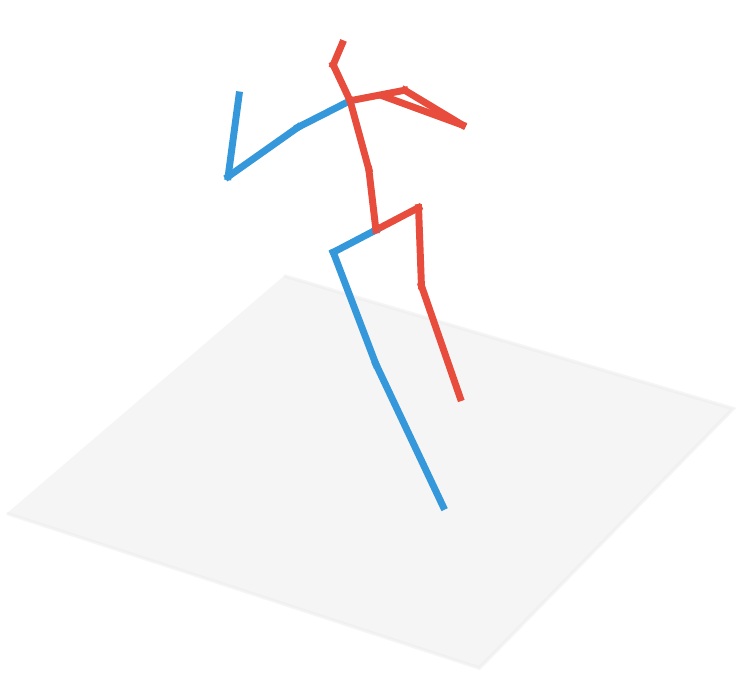}\hspace{0.9cm} &
\includegraphics[width = 0.12\linewidth]{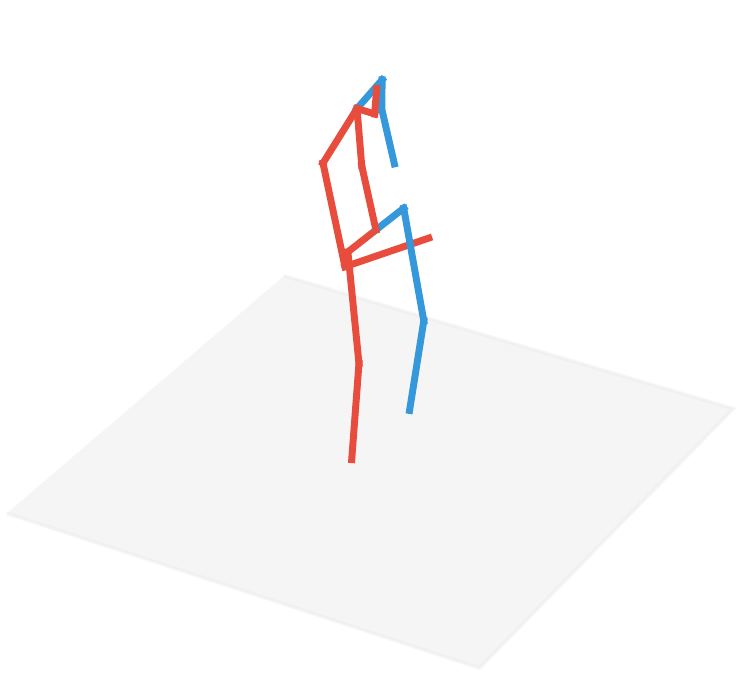} &
\includegraphics[width = 0.12\linewidth]{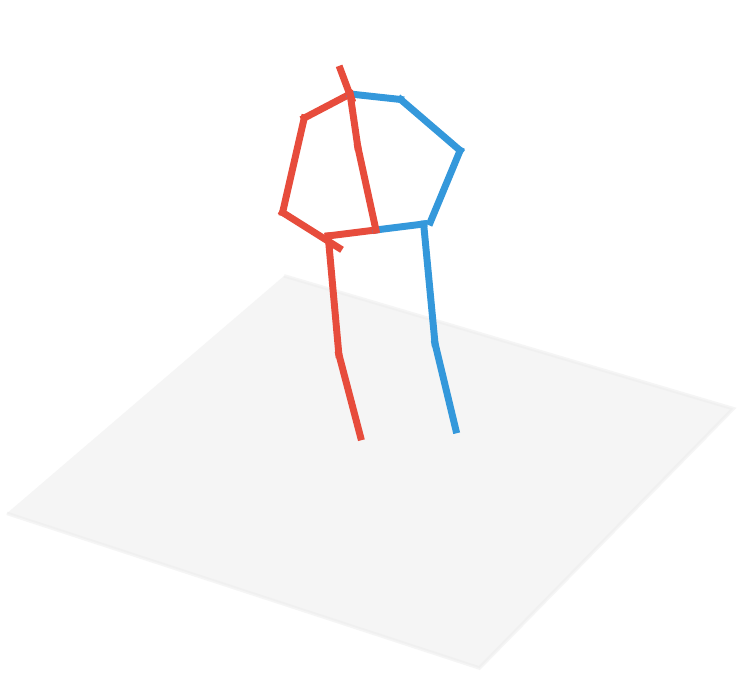} &
\includegraphics[width = 0.12\linewidth]{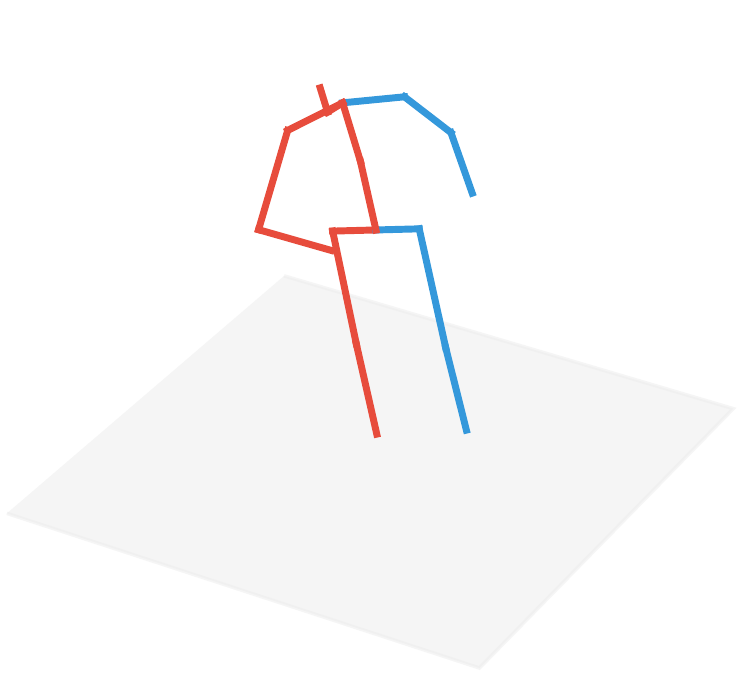} \\
\includegraphics[width = 0.12\linewidth]{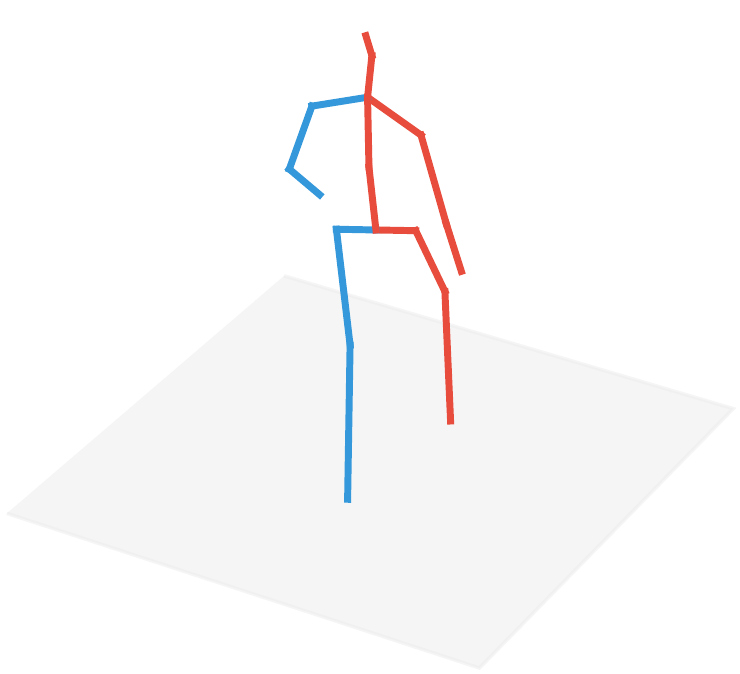} &
\includegraphics[width = 0.12\linewidth]{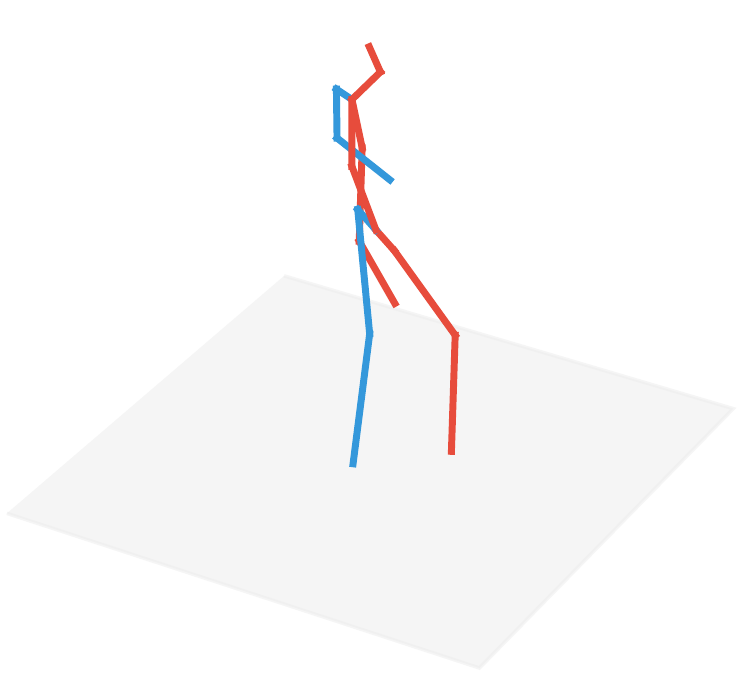} &
\includegraphics[width = 0.12\linewidth]{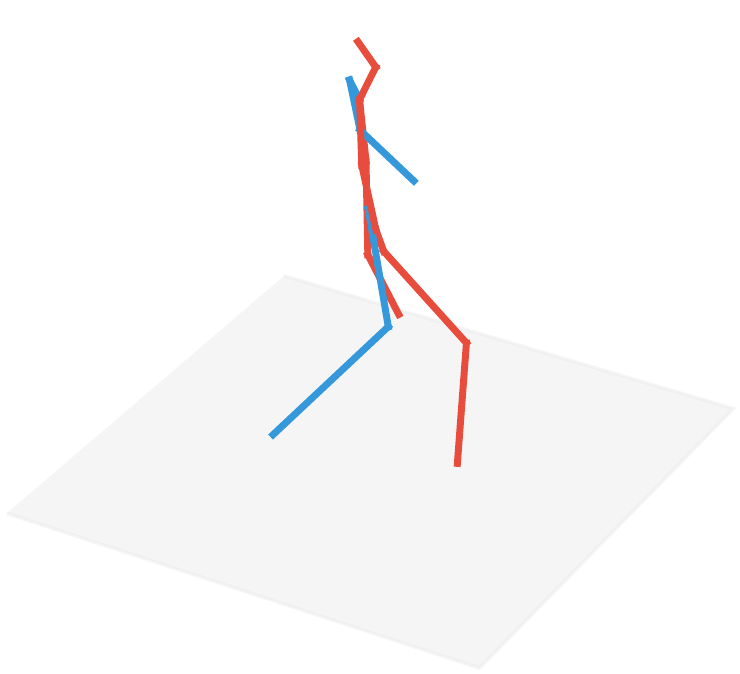}\hspace{0.9cm} &
\includegraphics[width = 0.12\linewidth]{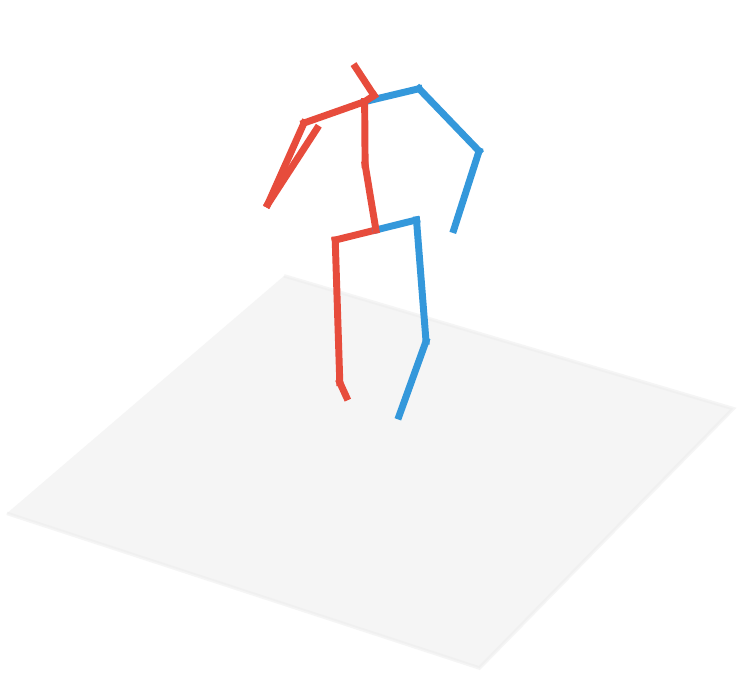} &
\includegraphics[width = 0.12\linewidth]{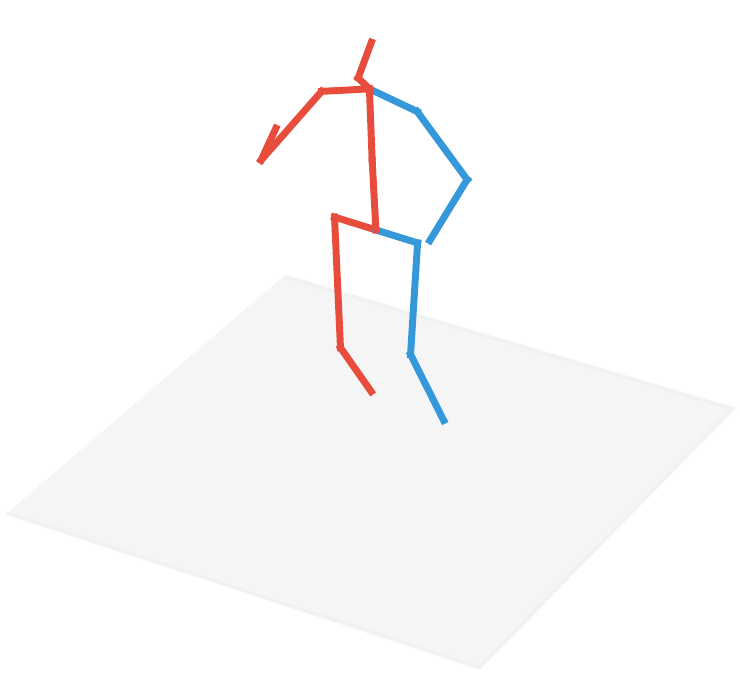} &
\includegraphics[width = 0.12\linewidth]{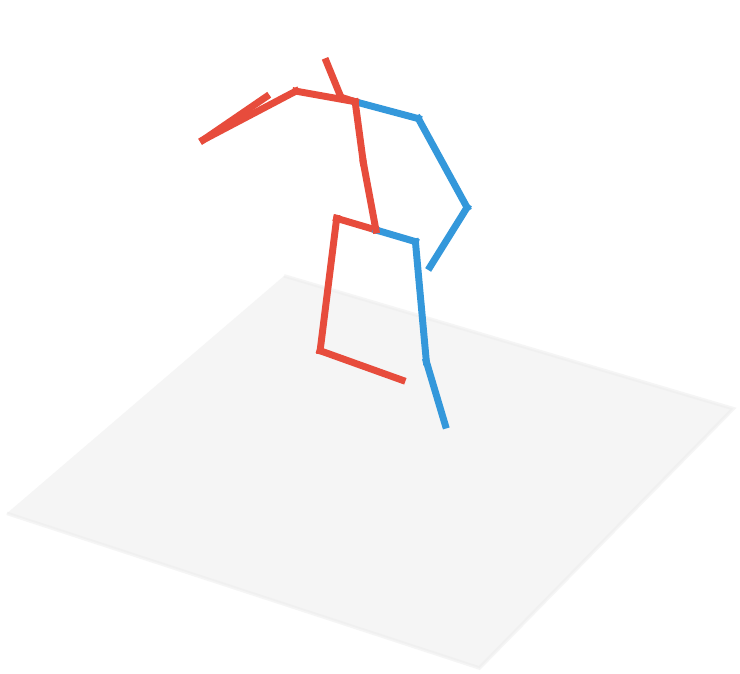} \\
\includegraphics[width = 0.12\linewidth]{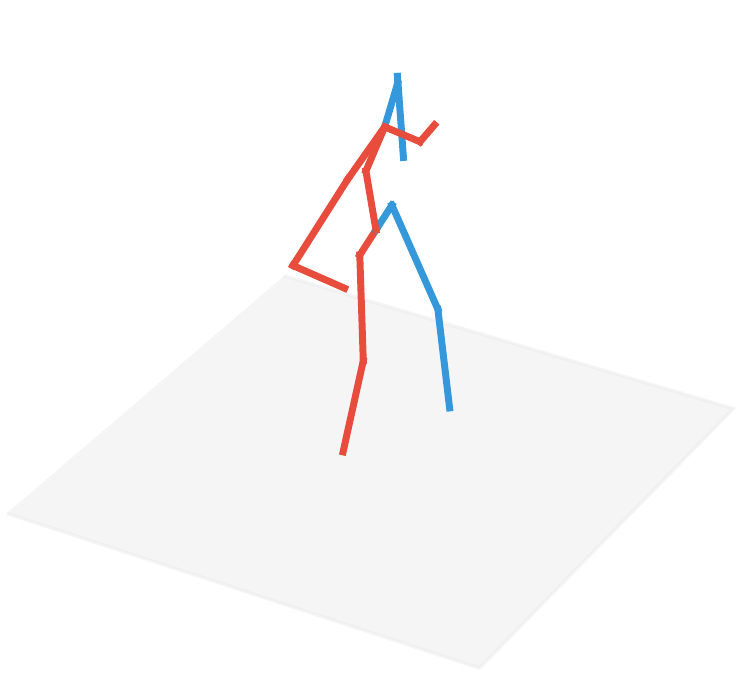} &
\includegraphics[width = 0.12\linewidth]{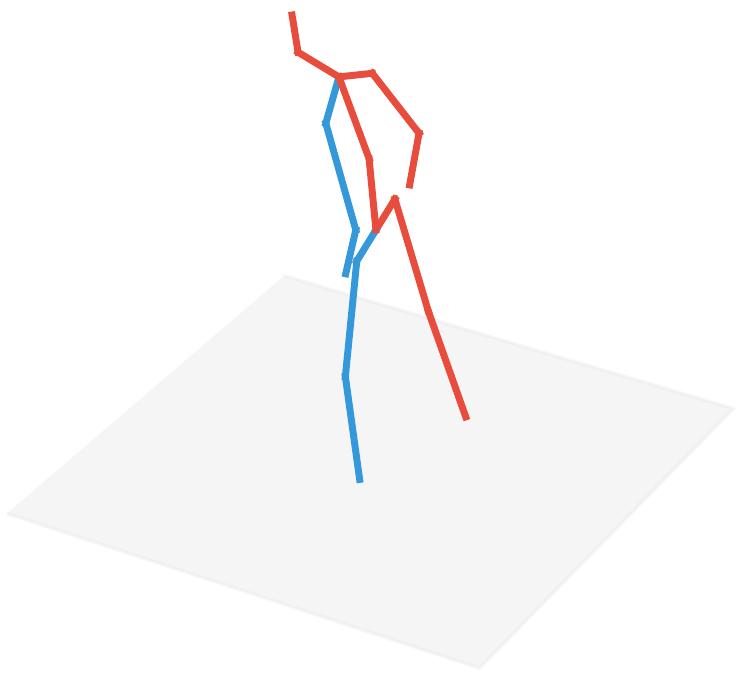} &
\includegraphics[width = 0.12\linewidth]{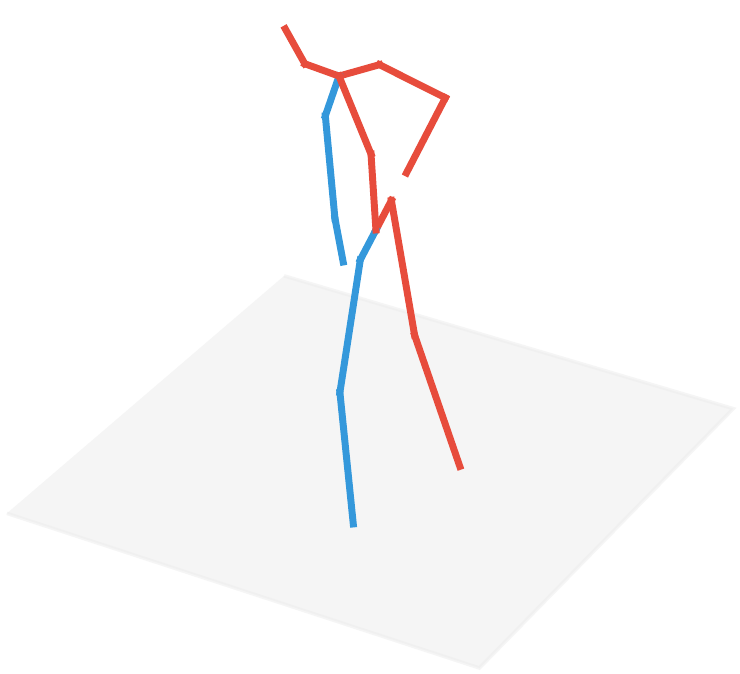}\hspace{0.9cm} &
\includegraphics[width = 0.12\linewidth]{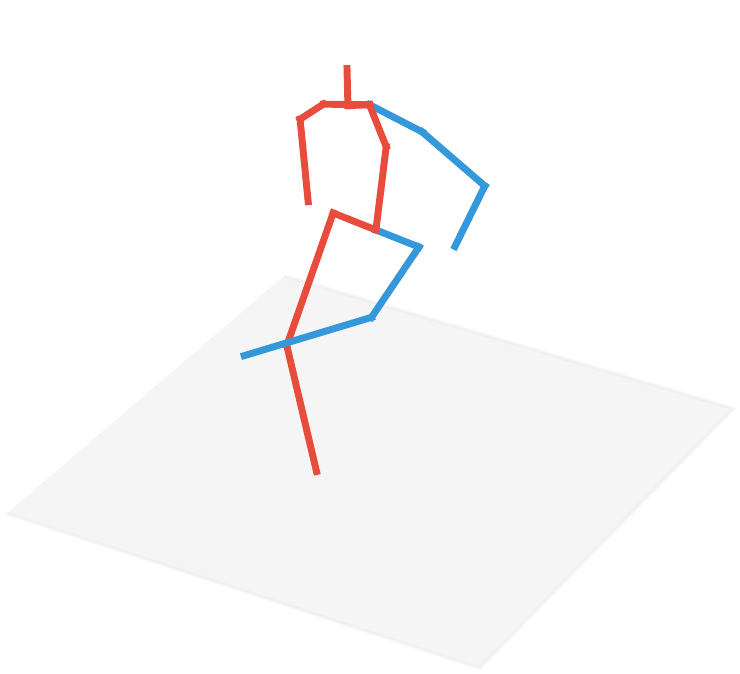} &
\includegraphics[width = 0.12\linewidth]{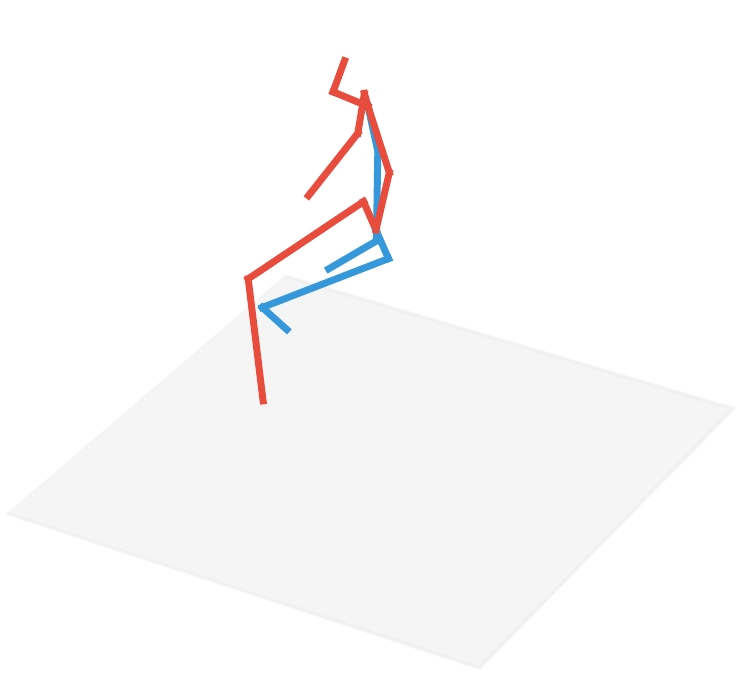} &
\includegraphics[width = 0.12\linewidth]{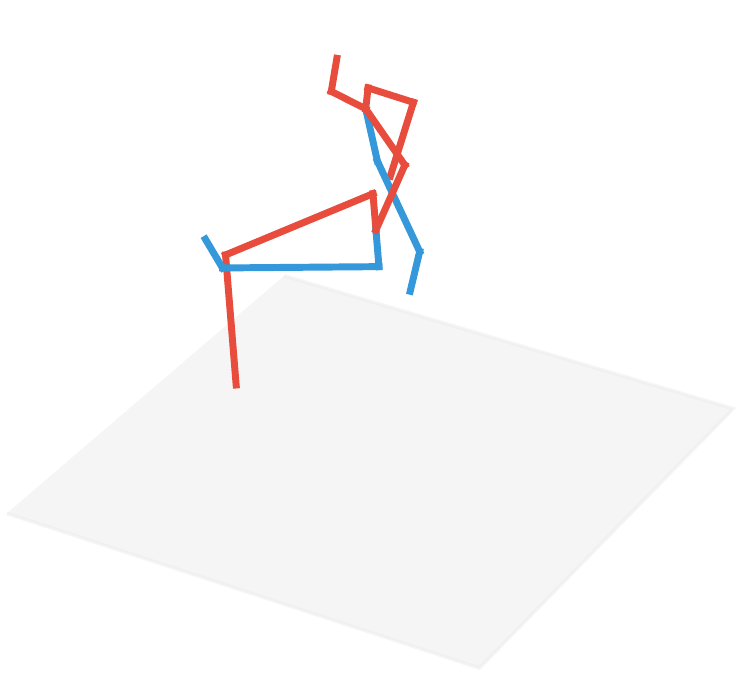} \\

\end{tabular}
\caption{\textbf{Rotation of the hidden embedding} There are $6 \times 3$ images. In each triplet of images, the first image is the input 3D skeleton $P_{3D}$, the second is the result after rotating the hidden layer ($g(Rh)$), the third is the ground truth 3D skeleton, but rotated ($RP_{3D}$). Bottom row, left triplet shows that even under large (180 degree) rotations the model produces high quality results. Bottom row, right triplet shows a failure case. }
\label{fig:embedding_rot}
\end{figure}

To show that the encoded hidden representation indeed behaves rotationally equivariant, we rotated the embedding, applied the decoder and compared the results to the expected rotated output. Formally, for an input 2D pose $P_{2D}$, we calculated both $g(Rf(P_{2D}))$ and $RP_{3D}$, where $R$ is a 3D rotation matrix and $P_{3D}$ is the ground truth 3D pose for $P_{2D}$. Results are presented in \autoref{fig:embedding_rot}. 

As seen in the figure, rotating the hidden embedding produces accurate predictions close to the ground truth even under large angles (bottom left pose on \autoref{fig:embedding_rot}). A failure case is shown on the bottom right of the figure. We found that sitting poses have higher errors probably because the database contains mostly standing poses.

\subsection{Ablation studies}

\begin{table}[ht]
\centering
\begin{subtable}[t]{.45\columnwidth}
\centering
    \begin{tabular}[t]{lc}
    \hline
    Variant & Error \\
    \hline
    Baseline* & 84.9 \\
    Baseline$^\dagger$ & 86.5 \\
    \hline
    w/o Siamese loss & 71.1  \\
    w/o Augmentation & 81.0 \\
    w/o Leaky ReLU & 67.1 \\
    \hline
    w/ All components & 65.8 \\
    \hline
    \end{tabular}
    \caption{Error of our method with components turned off. *Results from \cite{fang2018posegrammar}. $^\dagger$~Results of our implementation.}
\end{subtable}
\hspace{0.5cm}
\begin{subtable}[t]{.45\columnwidth}
\centering
    \begin{tabular}[t]{lc}
    \hline
    Variant & Error \\
    \hline
    No Augmentation & 86.5 \\
    Rot. Aug. & 76.6 \\
    Rot. Aug.+Noise & 69.6 \\
    \hline
    Ours w/o Aug & 81.0 \\
    \hline
    \end{tabular}
    \vspace{0.6cm}
    \caption{Error of the Baseline method with different augmentations}
\end{subtable}
\caption{\textbf{Ablation studies.} a) Mean joint error in millimeters with a single component of our method turned off. In the baseline algorithm  all components turned off. It is equivalent to the case of Martinez et al. \cite{3dbaseline}. b) The effect of different levels of augmentations on the baseline.} \label{tbl:ablation} 
\end{table}

We performed an ablation study to confirm the necessity of the components of our algorithm. If we remove all the components, our method is the same as the one in \cite{3dbaseline}. It is called \textit{Baseline} in \autoref{tbl:ablation}. Note that our implementation (marked with $^\dagger$ in the table) produces results slightly worse (1.6mm) than the one reported in \cite{fang2018posegrammar}. The table shows the performance of our method when turning off a single components.

Removing the siamese loss decreases the performance by 5.3mm, compared to all components turned on (71.1mm vs 65.8mm). 
Turning off augmentation decreases the performance the most among the components, however it is still better by 5.5mm (6.4\%) than the baseline algorithm. Finally, without Leaky ReLU the performance drops 1.3mm.

\begin{table}[ht]
\centering
    \begin{tabular}[t]{lc}
    \hline
    Variant & Error \\
    \hline
    PoseGrammar* & 72.8 \\
    PoseGrammar$^\dagger$  & 67.8 \\
    Siamese PoseGrammar & 65.0  \\ 
    \hline
    \end{tabular}
    \caption{\textbf{Using PoseGrammar as a base network.} Mean joint error for different PoseGrammar implementations. *Results published in \cite{fang2018posegrammar}. $^\dagger$Our implementation with more augmented viewpoints. }
    \label{tbl:siamese_posegrammar}
\end{table}

Furthermore, our equivariant embedding can be applied to other network structures, not only to Baseline. To show that the method is general, we also extended Fang et al.'s PoseGrammar \cite{fang2018posegrammar} network to have a siamese structure. 

The PoseGrammar network has a bottom part consisting of 4 residual blocks, and a top part built up from multiple bidirectional RNNs. The geometric embeddings together with the siamese loss were placed after the first and third residual blocks since the network has an intermediate supervision after the second and fourth blocks. Following the original training protocol, we first trained the bottom residual network only for 200 epochs and then finetuned the whole network with the RNN blocks on top for another 200 epochs. Results are presented in  \autoref{tbl:siamese_posegrammar}. First note that our implementation uses more augmented viewpoints than the original, already improving the error from 72.8mm to 67.8mm. Adding the siamese architecture with equivariant embedding further decreases the loss to 65mm.

\subsection{The effect of data augmentation}
\begin{figure}[ht]
    \centering
    \includegraphics[width=0.7\textwidth]{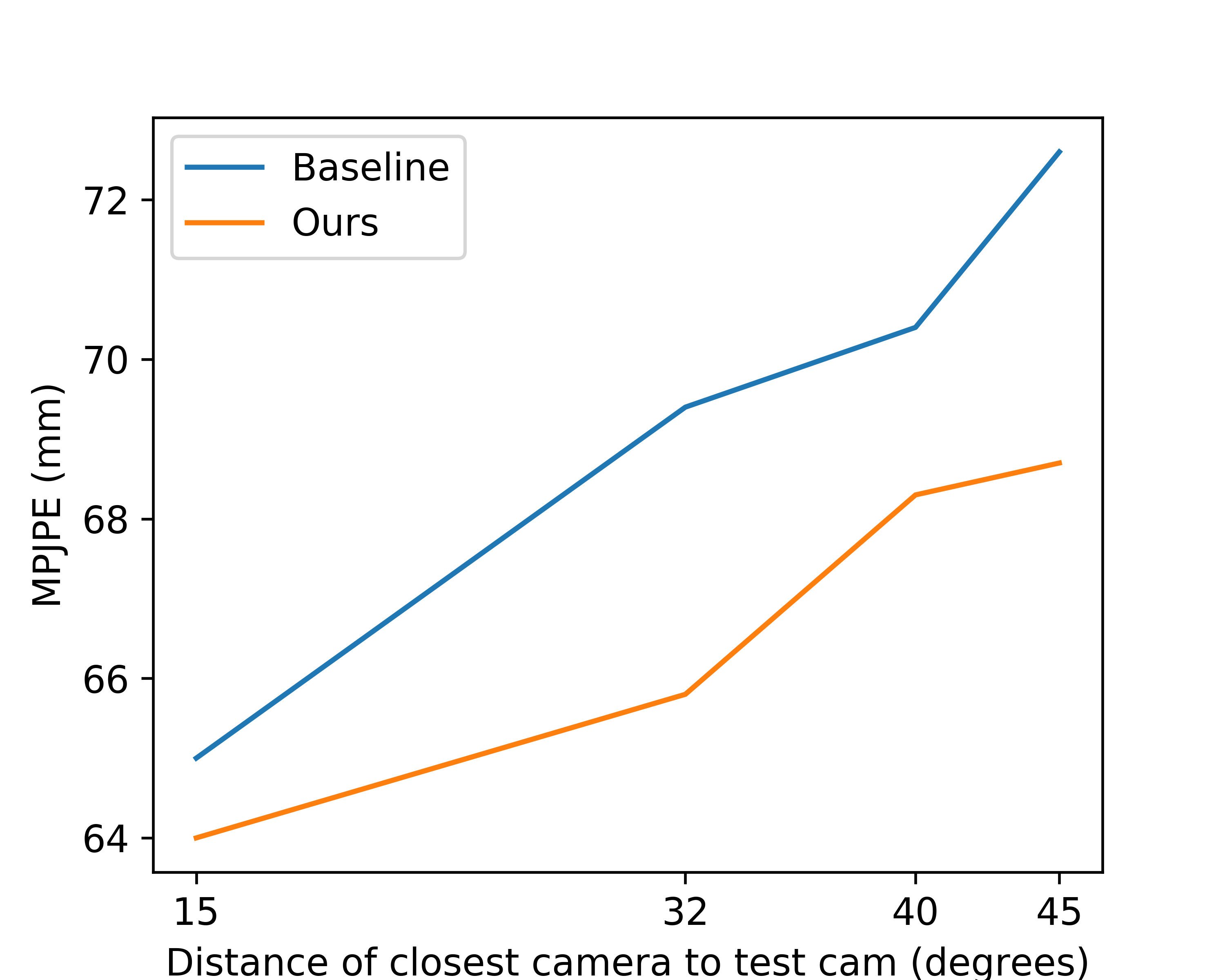}
    \caption{\textbf{The effect of synthetic camera placement on the model performance (Protocol \#3).} The x axis shows how far the closest camera is from the test camera in degrees. The closer a training camera is to the test camera the better the results are.}
    \label{fig:aug-effect}
\end{figure}

We also investigated how the siamese loss compares to augmentation. Note that the augmentation process has two steps: creating synthetic cameras by rotating existing ones around the subject and simulating the noise of the 2D pose estimator (for more details see \cite{fang2018posegrammar}). The siamese loss is only capable of replacing the camera rotation and not generating additional noise. Thus a lower bound on the performance of our architecture without augmentation is the performance of the baseline algorithm with only camera rotation augmentation (76.6mm). We achieve results that are halfway to the lower bound (81.1mm). 

Additionally, we analyzed how the number of synthetic cameras  affect the prediction performance. We found that it is not the number of cameras that the prediction performance depends on but the distance of the closest training cameras from the test camera. \autoref{fig:aug-effect} shows how the prediction performance changes as we create cameras closer to the test cam. In all cases, our method is better than the baseline. As we get closer to the test cam the gap in MPJPE decreases. Our method can achieve the same level of performance as the baseline with less augmentation.

\section{Conclusion and Future Work}
We have introduced a siamese network with an equivariant embedding that provides regularization for cross-camera 3D human pose estimation. It was shown that the method performs state-of-the-art if only 2D pose detection information is used. This distinction is important, as our method is orthogonal to others using image information (e.g. \cite{integralPose,drpose}) and can be integrated with those easily.

There are promising ways for improvements. One option is to go beyond 2D keypoint coordinates and use other information derived from the image, such as a pairwise ranking matrix \cite{drpose}. Other avenues not yet investigated include changing the siamese loss to a triplet loss and/or improvments in the input sampling. Both were found to have large effect on network performance \cite{facenet,amazon_siamese-sampling}.

\section*{Acknowledgements}
M.V. takes part in the ELTE Institutional Excellence Program (1783-3/2018/ FEKUTSRAT) supported by the Hungarian Ministry of Human Capacities. V.V. has been supported by the European Union, co-financed by the European Social Fund (EFOP-3.6.3-VEKOP-16-2017-00002). 

\textit{Author Contributions} M\'arton V\'eges developed the main thesis and performed most of the analysis. Viktor Varga ran the analysis on the MPII-2D database. Andr\'as L\H{o}rincz was the supervisor of the project.

\section*{References}

\bibliography{refs}

\begin{thebibliography}{10}
\expandafter\ifx\csname url\endcsname\relax
  \def\url#1{\texttt{#1}}\fi
\expandafter\ifx\csname urlprefix\endcsname\relax\def\urlprefix{URL }\fi
\expandafter\ifx\csname href\endcsname\relax
  \def\href#1#2{#2} \def\path#1{#1}\fi

\bibitem{openpose}
Z.~Cao, T.~Simon, S.-E. Wei, Y.~Sheikh, Realtime multi-person 2d pose
  estimation using part affinity fields, in: The IEEE Conference on Computer
  Vision and Pattern Recognition, 2017, pp. 1302--1310.

\bibitem{stacked_hourglass}
A.~Newell, K.~Yang, J.~Deng, Stacked hourglass networks for human pose
  estimation, in: European Conference on Computer Vision, 2016, pp. 483--499.

\bibitem{alphapose}
H.-S. Fang, S.~Xie, Y.-W. Tai, C.~Lu, Rmpe: Regional multi-person pose
  estimation, in: 2017 IEEE International Conference on Computer Vision (ICCV),
  2017, pp. 2353--2362.

\bibitem{3dbaseline}
J.~Martinez, R.~Hossain, J.~Romero, J.~J. Little, A simple yet effective
  baseline for 3d human pose estimation, in: The IEEE International Conference
  on Computer Vision, 2017, pp. 2659--2668.

\bibitem{fang2018posegrammar}
H.-S. Fang, Y.~Xu, W.~Wang, X.~Liu, S.-C. Zhu, Learning pose grammar to encode
  human body configuration for 3d pose estimation, AAAI.

\bibitem{h36m}
C.~Ionescu, D.~Papava, V.~Olaru, C.~Sminchisescu, Human3.6m: Large scale
  datasets and predictive methods for 3d human sensing in natural environments,
  IEEE Transactions on Pattern Analysis and Machine Intelligence 36~(7) (2014)
  1325--1339.

\bibitem{siameseSignature}
J.~Bromley, I.~Guyon, Y.~LeCun, E.~S\"{a}ckinger, R.~Shah, Signature
  verification using a "siamese" time delay neural network, in: Proceedings of
  the 6th International Conference on Neural Information Processing Systems,
  1993, pp. 737--744.

\bibitem{gorog}
G.~Pavlakos, X.~Zhou, K.~G. Derpanis, K.~Daniilidis, Coarse-to-fine volumetric
  prediction for single-image 3d human pose, in: The IEEE Conference on
  Computer Vision and Pattern Recognition, IEEE, 2017, pp. 1263--1272.

\bibitem{zhou2017}
X.~Zhou, Q.~Huang, X.~Sun, X.~Xue, Y.~Wei, Towards 3d human pose estimation in
  the wild: A weakly-supervised approach, in: The IEEE International Conference
  on Computer Vision (ICCV), 2017, pp. 398--407.

\bibitem{Sun2017compositional}
X.~Sun, J.~Shang, S.~Liang, Y.~Wei, Compositional human pose regression, The
  IEEE International Conference on Computer Vision (2017) 2621--2630.

\bibitem{Hossain2017temporal}
M.~R.~I. Hossain, J.~J. Little, Exploiting temporal information for 3d pose
  estimation (2017).
\newblock \href {http://arxiv.org/abs/1711.08585} {\path{arXiv:1711.08585}}.

\bibitem{integralPose}
X.~Sun, B.~Xiao, S.~Liang, Y.~Wei, Integral human pose regression, in: The
  European Conference on Computer Vision, 2018, pp. 529--545.

\bibitem{Luvizon2018softargmax}
D.~C. Luvizon, D.~Picard, H.~Tabia, 2d/3d pose estimation and action
  recognition using multitask deep learning, in: The IEEE Conference on
  Computer Vision and Pattern Recognition, 2018, pp. 5137--5146.

\bibitem{mpii-hp}
M.~Andriluka, L.~Pishchulin, P.~Gehler, B.~Schiele, 2d human pose estimation:
  New benchmark and state of the art analysis, in: The IEEE Conference on
  Computer Vision and Pattern Recognition, 2014, pp. 3686--3693.

\bibitem{lsp}
S.~Johnson, M.~Everingham, Clustered pose and nonlinear appearance models for
  human pose estimation, in: Proceedings of the British Machine Vision
  Conference, 2010, pp. 12.1--12.11.

\bibitem{pavlakos2018ordinal}
G.~Pavlakos, X.~Zhou, K.~Daniilidis, Ordinal depth supervision for 3d human
  pose estimation, in: The IEEE Conference on Computer Vision and Pattern
  Recognition, 2018, pp. 7307--7316.

\bibitem{fbipose}
Y.~Shi, X.~Han, N.~Jiang, K.~Zhou, K.~Jia, J.~Lu, Fbi-pose: Towards bridging
  the gap between 2d images and 3d human poses using forward-or-backward
  information.
\newblock \href {http://arxiv.org/abs/1806.09241} {\path{arXiv:1806.09241}}.

\bibitem{drpose}
M.~Wang, X.~Chen, W.~Liu, C.~Qian, L.~Lin, L.~Ma, Drpose3d: Depth ranking in 3d
  human pose estimation, in: Proceedings of the Twenty-Seventh International
  Joint Conference on Artificial Intelligen, 2018, pp. 978--984.

\bibitem{ronchi2018allrelative}
M.~R. Ronchi, O.~{Mac Aodha}, R.~Eng, P.~Perona, It's all relative: Monocular
  3d human pose estimation from weakly supervised data, in: Proceedings of the
  British Machine Vision Conference, 2018.

\bibitem{facenet}
F.~Schroff, D.~Kalenichenko, J.~Philbin, Facenet: A unified embedding for face
  recognition and clustering, Proceedings of the IEEE conference on computer
  vision and pattern recognition (2015) 815--823.

\bibitem{deepid2}
Y.~Sun, Y.~Chen, X.~Wang, X.~Tang, Deep learning face representation by joint
  identification-verification, in: Proceedings of the 27th International
  Conference on Neural Information Processing Systems, 2014, pp. 1988--1996.

\bibitem{doumanoglouSiamesePose}
A.~Doumanoglou, V.~Balntas, R.~Kouskouridas, T.-K. Kim, Siamese regression
  networks with efficient mid-level feature extraction for 3d object pose
  estimation (2016).
\newblock \href {http://arxiv.org/abs/1607.02257} {\path{arXiv:1607.02257}}.

\bibitem{siamese_headpose}
M.~Venturelli, G.~Borghi, R.~Vezzani, R.~Cucchiara, From depth data to head
  pose estimation: a siamese approach (2017).
\newblock \href {http://arxiv.org/abs/1703.03624} {\path{arXiv:1703.03624}}.

\bibitem{worrall2016harmonic}
D.~E. Worrall, S.~J. Garbin, D.~Turmukhambetov, G.~J. Brostow, Harmonic
  networks: Deep translation and rotation equivariance, in: The IEEE Conference
  on Computer Vision and Pattern Recognition, 2017, pp. 7168--7177.

\bibitem{rotatedMNIST}
H.~Larochelle, D.~Erhan, A.~Courville, J.~Bergstra, Y.~Bengio, An empirical
  evaluation of deep architectures on problems with many factors of variation,
  in: Proceedings of the 24th International Conference on Machine Learning,
  2007, pp. 473--480.

\bibitem{cohen2018spherical}
T.~Cohen, M.~Geiger, J.~K{\"{o}}hler, M.~Welling, Spherical cnns, in:
  Proceedings of the 6th International Conference on Learning Representation,
  2018, pp. 1--15.

\bibitem{esteves2018so3equivariance}
C.~Esteves, C.~Allen-blanchette, A.~Makadia, K.~Daniilidis, Learning so(3)
  equivariant representations with spherical cnns, in: Proceedings of the
  European Conference on Computer Vision, 2018, pp. 52--68.

\bibitem{modelnet40}
Z.~Wu, S.~Song, A.~Khosla, F.~Yu, L.~Zhang, X.~Tang, J.~Xiao, 3d shapenets: A
  deep representation for volumetric shapes, in: The IEEE Conference on
  Computer Vision and Pattern Recognition, 2015, pp. 1912--1920.

\bibitem{shrec17}
M.~Savva, F.~Yu, H.~Su, A.~Kanezaki, T.~Furuya, R.~Ohbuchi, Z.~Zhou, R.~Yu,
  S.~Bai, X.~Bai, M.~Aono, A.~Tatsuma, S.~Thermos, A.~Axenopoulos, G.~T.
  Papadopoulos, P.~Daras, X.~Deng, Z.~Lian, B.~Li, H.~Johan, Y.~Lu, S.~Mk,
  Shrec'17 track: Large-scale 3d shape retrieval from shapenet core55, in:
  Eurographics Workshop on 3D Object Retrieval, 2017.

\bibitem{helge_geometry-aware}
H.~Rhodin, M.~Salzmann, P.~Fua, Unsupervised geometry-aware representation for
  3d human pose estimation, in: European Conference on Computer Vision, 2018,
  pp. 750--767.

\bibitem{worall2017interpretable}
D.~E. Worrall, S.~J. Garbin, D.~Turmukhambetov, G.~J. Brostow, Interpretable
  transformations with encoder-decoder networks, The IEEE International
  Conference on Computer Vision (2017) 5737--5746.

\bibitem{batchnorm}
S.~Ioffe, C.~Szegedy, Batch normalization: Accelerating deep network training
  by reducing internal covariate shift, in: International Conference on Machine
  Learning, 2015, pp. 448--456.

\bibitem{dropout}
N.~Srivastava, G.~Hinton, A.~Krizhevsky, I.~Sutskever, R.~Salakhutdinov,
  Dropout: A simple way to prevent neural networks from overfitting, The
  Journal of Machine Learning Research 15~(1) (2014) 1929--1958.

\bibitem{amazon_siamese-sampling}
C.~Wu, R.~Manmatha, A.~J. Smola, P.~Kr{\"{a}}henb{\"{u}}hl, Sampling matters in
  deep embedding learning, in: The IEEE International Conference on Computer
  Vision, 2017, pp. 2859--2867.

\end{thebibliography}

\end{document}